\documentclass[11pt]{article}

\usepackage[preprint]{acl}

\usepackage{times}
\usepackage{latexsym}

\usepackage[T1]{fontenc}

\usepackage[utf8]{inputenc}

\usepackage{microtype}

\usepackage{inconsolata}

\usepackage{graphicx}

\usepackage{microtype}
\usepackage{graphicx}
\usepackage{subcaption}
\usepackage{booktabs} 
\usepackage{tabularx}
\usepackage{multirow}
\usepackage{longtable}
\usepackage{float}
\usepackage{supertabular}

\usepackage{xcolor}
\usepackage{tcolorbox}
\tcbuselibrary{listings, breakable, skins}

\usepackage{tcolorbox}
\newtcblisting{promptbox}[1][]{
  enhanced,
  breakable,
  colback=gray!5,
  colframe=gray!60,
  boxrule=0.5pt,
  arc=2pt,
  left=6pt, right=6pt, top=6pt, bottom=6pt,
  listing only,
  listing options={
    basicstyle=\ttfamily\small,
    breaklines=true,
    keepspaces=true
  },
  title=#1
}

\usepackage{hyperref}

\usepackage{amsmath}
\usepackage{amssymb}
\usepackage{mathtools}
\usepackage{amsthm}

\usepackage[capitalize,noabbrev]{cleveref}

\usepackage[linesnumbered,ruled,vlined]{algorithm2e}
\usepackage{amsmath}
\usepackage{algorithmic}

\title{JFTA-Bench: Evaluate LLM's Ability \\
  of Tracking and Analyzing Malfunctions Using Fault Trees}
\author{
  \textbf{Yuhui Wang}$^{1,2}$\thanks{Equal contribution.},
  \textbf{Zhixiong Yang}$^{1}$\footnotemark[1],
  \textbf{Ming Zhang}$^{1}$\footnotemark[1],
  \textbf{Shihan Dou}$^{1}$,
  \textbf{Zhiheng Xi}$^{1}$,
  \textbf{Di Liang}$^{2}$,
  \\
  \textbf{Enyu Zhou}$^{1}$,
  \textbf{Senjie Jin}$^{1}$,
  \textbf{Yujiong Shen}$^{1}$,
  \textbf{Dingwei Zhu}$^{1}$,
  \textbf{Yi Dong}$^{1}$,
  \textbf{Yujia Chen}$^{1}$,
  \\
  \textbf{Haibo Shi}$^{2}$,  
  \textbf{Tao Gui}$^{1,3,4}$,
  \textbf{Qi Zhang}$^{1,3,4}$,
  \textbf{Xuanjing Huang}$^{1,3,4}$
  \\
  $^1$Fudan University \\
  $^2$Tencent \\
  $^3$Shanghai Artificial Intelligence Laboratory \\
  $^4$Shanghai Key Laboratory of Intelligent Information Processing\\
  \small{\texttt{\{qz,tgui\}@fudan.edu.cn}}
}
\begin{document}
\maketitle
\begin{abstract}
In the maintenance of complex systems, fault trees are used to locate problems and provide targeted solutions. To enable fault trees stored as images to be directly processed by large language models, which can assist in tracking and analyzing malfunctions, we propose a novel textual representation of fault trees.  Building on it, we construct a benchmark for multi-turn dialogue systems that emphasizes robust interaction in complex environments, evaluating a model's ability to assist in malfunction localization, which contains $3130$ entries and $40.75$ turns per entry on average. We train an end-to-end model to generate vague information to reflect user behavior and introduce long-range rollback and recovery procedures to simulate user error scenarios, enabling assessment of a model's integrated capabilities in task tracking and  error recovery, and Gemini 2.5 pro achieves the best performance.
\end{abstract}

\begin{figure*}[ht]
  \vskip 0.2in
  \begin{center}
    \centerline{\includegraphics[width=\textwidth]{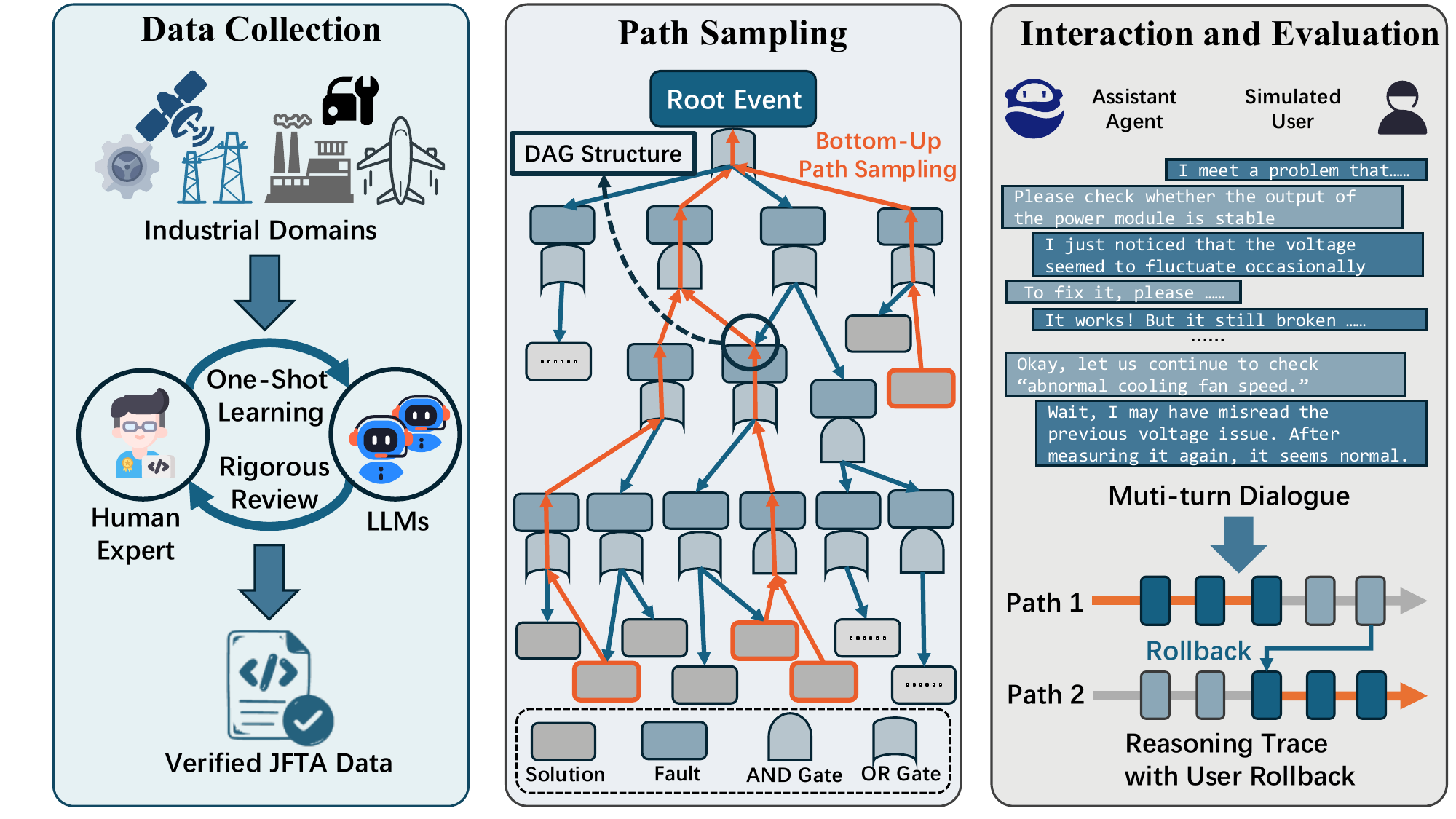}}
    \caption{
      The left panel illustrates the Human-in-the-loop data collection process. The middle panel depicts the path sampling procedure capable of handling DAG structures. The right panel demonstrates the multi-turn evaluation framework featuring user rollback scenarios.
    }
    \label{main-fig}
  \end{center}
\end{figure*}

\section{Introduction}

Fault Tree Analysis (FTA) is a top-down, deductive failure analysis methodology that has been widely adopted in the maintenance of complex systems for fault localization and decision support~\cite{lee2009fault}. Starting from a predefined system-level top event, FTA models the causal relationships between failures using Boolean logic and progressively decomposes system failures into basic events such as component-level malfunctions~\cite{ruijters2015fault}. Subsequent extensions, including Dynamic Fault Trees~\cite{vcepin2002dynamic}, further enrich the expressive power of fault trees by introducing additional logic gates and temporal dependencies, enabling more fine-grained reasoning in realistic industrial scenarios.

In parallel, large language models (LLMs), represented by systems such as GPT-5~\cite{singh2025openai}, Gemini 3 pro~\cite{gemini3pro2025modelcard} and DeepSeek-V3.2~\cite{deepseekai2025deepseekv32}, have achieved rapid advances in recent years. With strong capabilities in in-context learning~\cite{dong2024survey},  recent research on LLM-based agents~\cite{luo2025large} further highlights their applicability in complex decision-making and interactive problem-solving tasks, suggesting promising opportunities for assisting fault tree–based diagnosis. To support both professional technicians and non-expert users in efficiently locating malfunctions in complex systems without requiring manual exploration of extensive fault trees, we leverage LLMs to assist and streamline the diagnosis workflow.

However, existing fault tree standards and datasets are predominantly image-based~\cite{vesely1981fault}, whereas most open-source and commercial LLMs are pretrained primarily on textual corpora~\cite{zhao2023survey}. Although multimodal models attempt to bridge this gap by projecting visual features into textual embedding spaces via connector-based alignment~\cite{liu2023visual}, such approaches inevitably introduce information loss and limit precise structural reasoning~\cite{feng2025vision,liu2023deplot}. 

To address this mismatch, we propose JFTA (JSON-based Fault Tree Analysis), a structured, extensible, and model-friendly textual representation of fault trees. By organizing different logic gates through explicit structural compositions and enabling cross-branch references, our representation supports faithful textual encoding of complex fault trees beyond simple tree structures. It strikes a balance between natural language expressiveness and formal structural rigor.

Building on this representation, we construct a benchmark to evaluate the ability of LLMs to learn and reason over fault trees. Figure \ref{main-fig} provides an overview of the JFTA-Bench construction and evaluation pipeline. We collect and process a Chinese-language dataset of 126 fault trees from diverse domains, with an average of 140 nodes per tree.  From these fault trees, we sample 3,130 distinct fault path groups and organize them into multiple difficulty levels. 
To further approximate real-world conditions, we explicitly model user errors during interaction. We introduce long-range state rollback and recovery procedures, where users may correct or revise earlier statements in the dialogue history. This design enables systematic evaluation of a model's integrated capabilities in task tracking, consistency maintenance, and error recovery across extended multi-turn conversations.

The benchmark evaluates whether a model can correctly localize all root causes and propose appropriate solutions within a limited number of interaction turns. According to our evaluation, Gemini 2.5 pro~\cite{comanici2025gemini25pushingfrontier} achieves the best overall performance, successfully passing 53.76\% of the test cases by correctly identifying all malfunctions. Among open-source models, DeepSeek-V3.2 performs the strongest, reaching a success rate of 41.40\%. Further analysis indicates that model failures are primarily attributable to deficiencies in planning capability.

To simulate realistic and consistent user behavior in complex interactive settings, we select 100 fault trees from the dataset to train a user simulator that supports multi-turn dialogue evaluation. The trained user generates vague, uncertain, and occasionally biased information, better reflecting real user characteristics and potential adversarial behaviors. The remaining 26 fault trees are used for testing, on which the user simulator achieves a correctness rate of 99.98\%.

In summary, our contributions are threefold:
\begin{itemize}
\item We propose JFTA, a structured, extensible, and easy-to-understand textual representation for fault trees.

\item We construct JFTA-Bench, a benchmark for evaluating fault localization and error recovery capabilities of LLMs in multi-turn dialogue settings.

\item We train a user simulator that generates vague, uncertain, and biased responses to support realistic multi-turn interaction evaluation.
\end{itemize}

\section{Related Work}

\subsection{Fault Tree Analysis}
Fault Tree Analysis (FTA) was originally introduced by Bell Laboratories in 1962 as a qualitative design aid for system reliability analysis~\cite{ruijters2015fault}. It was later adopted and formalized by Boeing in the aerospace industry~\cite{eckberg1963ws}, where it evolved into a practical analytical and computational tool for diagnosing and maintaining complex engineered systems~\cite{ericson1999fault}. Since then, FTA has been widely applied across safety-critical domains such as aviation~\cite{goldberg1994system}, chemical process~\cite{american1985guidelines}, and so on. Classical FTA represents causal relationships between failures using Boolean logic, primarily through AND and OR gates, and decomposes a predefined system-level top event into basic events corresponding to component-level failures~\cite{vesely1981fault}. Dynamic Fault Trees~\cite{vcepin2002dynamic} introduce additional logic gates and temporal dependencies, and the International Electrotechnical Commission established the IEC 61025 standard~\cite{iec2006fault}, which defines FTA as a globally accepted formalism for reliability and safety analysis.

\subsection{Evaluation of Multi-Turn Dialogue Systems}

Existing research commonly categorizes dialogue systems into task-oriented dialogue (TOD) systems and open-domain dialogue (ODD) systems~\cite{wang2023survey}. Our work can be viewed as an extension of TOD evaluation into the domain of fault tree–based diagnosis, which exhibits inherently multi-turn, highly structured, and domain-specific characteristics. For TOD systems, a number of widely adopted benchmarks have been proposed, including general-purpose datasets such as MultiWOZ~\cite{budzianowski2018multiwoz} and CrossWOZ~\cite{zhu2020crosswoz}, as well as domain-specific benchmarks such as KddRES~\cite{wang2020kddres}, Transfer-TOD~\cite{zhang2024transfertod}, and PFDial~\cite{zhang2025pfdial}. Compared with existing multi-turn evaluation efforts, such as MT-Eval~\cite{kwan2024mt}, which evaluates dialogue performance across four domains but with relatively short interaction lengths, or MT-Bench-101~\cite{bai2024mt}, which further introduces a hierarchical evaluation framework for multi-turn dialogue, our benchmark emphasizes long-horizon interactions involving complex reasoning over structured fault trees, and additionally incorporates a trained user simulator to generate realistic and imperfect user behaviors. 

\subsection{Evaluation of Agent Capabilities}

Recent studies on LLM-based agents have proposed a variety of benchmarks that emphasize multi-step reasoning, tool use, memory, and self-reflection capabilities~\cite{yehudai2025survey}. For example, $\tau$-Bench~\cite{yao2024tau} evaluates an agent's ability to simulate customer service interactions in aviation and retail domains, while CRMArena~\cite{huang2025crmarena} assesses agent performance in customer relationship management scenarios that require executing multi-step operations under domain-specific policies. MINT~\cite{wang2023mint} evaluates planning capabilities in interactive environments, focusing on how agents adapt their strategies based on environmental feedback. In many agent benchmarks, tools are treated as interfaces with parameterized actions in uncertain environments~\cite{patilberkeley}. In contrast, fault trees define a set of valid structural states in advance, and the corresponding remediation actions can be determined once the underlying failures are correctly identified. Consequently, our work places less emphasis on tool-use ability.

\section{The JFTA Representation}
\label{sec:JFTA}

Existing fault tree representations are predominantly diagrammatic or semi-formal, typically distributed as static images~\cite{vesely1981fault}. While effective for human interpretation, such formats are less suited for direct processing by large language models~\cite{tang2025basic}. To bridge this gap, we propose a structured textual syntax based on JSON, termed JFTA.  JFTA preserves the core deductive logic of FTA while providing a formal, parseable, and extensible format, and strikes a balance between natural language expressiveness and formal structural rigor. On one hand, node names and solution descriptions are presented in natural language, facilitating semantic understanding. On the other hand, the underlying architecture enforces strict syntactic and logical constraints, supporting programmatic verification. We provide detailed information and example of JFTA in Appendix \ref{apd:JFTA}.


\subsection{Syntax Definition}

In JFTA, a fault tree is defined as a rooted structure that can represent a Directed Acyclic Graph (DAG). Each node constitutes a semantic unit relevant to the fault, representing a different-level event, or a logic gate.  JFTA supports a flexible set of logical and semantic nodes.



\textbf{Logic Gate Nodes:} These nodes formally describe the causal relationships between parent and child events. JFTA incorporates standard Boolean operators, including AND, OR, and XOR gates, to model common failure modes. The syntax is highly extensible, allowing for the integration of additional logic gates through custom definitions.

\textbf{Semantic Nodes:} These nodes distinguish between intermediate states that require further decomposition and terminal nodes that represent the end of the analysis. Specifically, intermediate nodes represent fault states that must be further analyzed, while leaf nodes are designated as solution nodes, containing natural language descriptions of specific repair actions or operational procedures.


\subsection{Hierarchical Structure and References}

The hierarchy between nodes is primarily expressed through a recursive nested structure, where a parent node contains a list of child nodes. However, complex systems often exhibit shared failure modes where a single subsystem failure contributes to multiple higher-level events. To avoid redundancy and explosion in tree size, JFTA introduces an explicit linking mechanism. Nodes can reference previously defined nodes via their unique IDs. This transforms the representation from a strict tree into a DAG, enhancing compactness and maintainability while preserving semantic consistency.

\section{JFTA-Bench}

To evaluate the capacity LLMs to reason over fault trees via in-context learning, we construct JFTA-Bench. Built upon the JFTA syntax defined in Section \ref{sec:JFTA}, this benchmark assesses a model's ability to assist users in malfunction localization within multi-turn dialogue settings. It emphasizes understanding complex context, strategic planning, and maintaining robust interaction in dynamic environments.

Figure \ref{main-fig} provides an overview of the JFTA-Bench construction and evaluation pipeline. The left part illustrates the fault tree collection process. The middle part shows how fault paths are extracted from fault trees in a bottom-up manner, capturing valid combinations of failures under complex logic gates. The right part depicts the interaction framework, where a simulated user engages with the LLM assistant under test, including scenarios with long-range state rollback and recovery.

\paragraph{Formal Definition}
We formalize each JFTA-Bench instance as : 
\[
x=(G, \mathcal{F}_1,\mathcal{P}_1,\pi_1,\mathcal{F}_2,\mathcal{P}_2,\pi_2,\rho).
\]

$G$ is a JFTA-formatted fault tree.
For $k\in\{1,2\}$, $\mathcal{F}_k$ denotes the set of ground-truth bottom-level faults, $\mathcal{P}_k$ is the corresponding set of root-to-fault paths, and $\pi_k$ is the ideal DFS-style traversal path induced by $\mathcal{P}_k$. 
The first group $(\mathcal{F}_1,\mathcal{P}_1,\pi_1)$ defines the initial diagnostic state, while the second group $(\mathcal{F}_2,\mathcal{P}_2,\pi_2)$ defines the corrected state after rollback.

$\rho=(v_{\mathrm{trig}}, v_{\mathrm{corr}})$ is defined as the rollback setting , where $v_{\mathrm{trig}}$ is the rollback trigger node on the first traversal path $\pi_1$. 
When the interaction reaches $v_{\mathrm{trig}}$, the user switches from the first fault setting to the second one. 
Let $\mathrm{LCP}(\pi_1,\pi_2)$ denote the longest common prefix of $\pi_1$ and $\pi_2$. 
The corrected-observation node $v_{\mathrm{corr}}$ is defined as the last node of this prefix:$v_{\mathrm{corr}}=\mathrm{Last}\bigl(\mathrm{LCP}(\pi_1,\pi_2)\bigr).$
It satisfies either $v_{\mathrm{corr}}\in V(\mathcal{P}_1)\setminus V(\mathcal{P}_2)$ or $v_{\mathrm{corr}}\in V(\mathcal{P}_2)\setminus V(\mathcal{P}_1)$ where $V(\mathcal{P}_k)=\bigcup_{p\in\mathcal{P}_k} p$, indicating that its observation must be revised after rollback. 

At each turn $t$, the assistant receives the fault tree and the dialogue history $H_{<t}$, and outputs either a fault query or a solution proposal. 
The user simulator receives the fault tree, the active path set $\mathcal{P}_k$, and the assistant action, where $k=1$ before rollback and $k=2$ after rollback. 
Let $\widehat{\mathcal{F}}$ be the set of fault nodes whose solutions are finally confirmed as effective. 
The instance is considered successful iff $\widehat{\mathcal{F}}=\mathcal{F}_2.$

\begin{table*}[htbp]
    \centering
    \begin{tabular}{ccccccc}
        \toprule
        Level & Size & Assistant Prompt Length & User Prompt Length & Interaction Turns & Error Count \\
        \midrule
        1 & 1103 & 5935.04 & 1096.87 & 20.17 & 1.50  \\
        2 & 1116 & 5924.64 & 1218.17 & 42.91 & 3.49  \\
        3 & 911  & 5939.75 & 1395.38 & 63.05 & 6.29 \\
        \bottomrule
    \end{tabular}
    \caption{Statistics of JFTA-Bench}
    \label{tab:statistics}
\end{table*}

\subsection{Fault Tree Collection}
\label{sec:FT-collection}
We first identified 24 domains requiring the maintenance of complex systems, such as Power \& Energy, Aerospace. Within these domains, we defined 126 specific failure scenarios (see Appendix \ref{apd:domains} for the full list).

To construct the dataset, we employed a human-in-the-loop generation pipeline. First, human experts manually annotated high-quality fault trees for three representative scenarios using strict JFTA syntax. These served as seed examples. Subsequently, we utilized GPT-4o~\cite{hurst2024gpt} and Claude Sonnet 4.5~\citep{anthropic2025claude45} in a One-Shot setting. By providing the complete JFTA syntax definition and the expert-annotated examples, we prompted the LLMs to generate fault trees for the remaining scenarios. The generated trees underwent rigorous review by human experts to ensure logical correctness and adherence to structural constraints, resulting in our final set of 126 verified fault trees.

\subsection{Path Extraction and Dataset Construction}

JFTA supports extensible logic gates and cross-branch references. Consequently, path sampling requires specific traversal logic. For instance, if a path traverses an AND gate, all child nodes of that gate must be included in the fault path. We extracted paths via a bottom-up approach, starting from sets of 1 to 6 initial basic failures. Using Algorithm \ref{alg:sampling}, we extended these failures upward to the root event. We classify test cases into three difficulty levels according to the number of final underlying failures after expansion: Level 1 for 1--2 failures, Level 2 for 3--4 failures, and Level 3 for 5 or more failures.

\subsection{Error Simulation and Recovery}
Real-world users often make observation errors and need to correct prior statements. To evaluate the model's robustness against such volatility, we introduced long-range state rollback and recovery scenarios.  This setup tests whether the model can accurately backtrack within the dialogue history, update its internal belief state, and proceed with the correct fault diagnosis based on the revised information.

\subsection{Interaction Framework}

We adopt the ReAct framework~\cite{yao2022react} for the evaluation. The model under test functions as an assistant interacting with a simulated User. In each turn, the assistant can either ask the user to verify the occurrence of a specific fault node or propose a solution for a confirmed failure. The user responds to the assistant's inquiries or confirms whether a proposed solution is effective.

To mimic the complexity of real-world diagnostics, user responses are not binary (True/False). Instead, the user provides vague or context-dependent descriptions based on the observation from environments or vague description, requiring the assistant to parse and infer the actual fault status. To achieve this, we trained a specialized user simulator capable of generating ambiguous yet consistent responses, detailed in Section \ref{sec:user-train}.

\subsection{Dataset Statistics}

Our dataset is derived from 126 complex fault trees, averaging 140 nodes per tree. We sampled $3130$ entries, where each entry contains two distinct paths to support the error recovery evaluation. On average, each session requires $40.75$ interaction turns to resolve. To support the ReAct interaction, we designed specific system prompts for both the user and assistant (see Appendix \ref{apd:prompt-user} and \ref{apd:prompt-assistant}). The objective is for the model, acting as an assistant, to start from an observed top-level symptom and progressively localize all faults and provide solutions within limited turns. Table \ref{tab:statistics} summarizes the key statistics of JFTA-Bench. The prompt length is calculated by the Qwen3-8B~\cite{yang2025qwen3} tokenizer.

\section{User Simulator Training}
\label{sec:user-train}

We trained a user simulator based on Qwen3-8B to simulate realistic user behaviors, including providing ambiguous feedback and maintaining consistency. The training pipeline consists of two stages: Supervised Fine-Tuning  for behavior cloning and Reinforcement Learning for response optimization.

\subsection{Data Collection and Construction.}
We utilized the 126 fault trees collected in Section \ref{sec:FT-collection}, partitioning them into a training set (100 trees) and a testing set (26 trees). For each fault tree , we constructed training samples by iterating through all possible fault nodes or solution nodes. We generated potential user responses  (Accept or Reject) and sampled corresponding path sets where paths either include or exclude consistent with the response.

To improve model robustness and prevent hacking behaviors (e.g., the user leaking the entire fault path or list of failures), we augmented the dataset with negative samples. We simulated invalid assistant requests, for instance, asking for unresolved faults, querying multiple faults simultaneously, or requesting irrelevant information, and paired them with rejection responses.

We employed an ensemble of advanced LLMs, including Gemini 2.5 flash~\cite{comanici2025gemini}, GPT-4o, and Claude Sonnet 4.5 randomly to synthesize dialogue data via One-Shot learning. Specialized prompts were designed to encourage vague expressions, mimicking non-expert users, which is shown in Appendix \ref{apd:prompt-user-data}. The target output was formatted in JSON, containing:

\texttt{action}: Classification of the assistant's intent.

\texttt{name}: The specific fault or solution queried.

\texttt{return}: The logical feedback (Accept/Reject).

\texttt{response}: The natural language response, specifically stylized to be ambiguous.

Detailed statistics of the collected user simulation dataset are provided in Appendix~\ref{apd:user-data-statistics}.

\subsection{Behavior Cloning}
We first fine-tuned the Qwen3-8B model using the LLaMA-Factory~\cite{zheng2024llamafactory} framework on the collected dataset to obtain the behavior-cloned model, $U_{bc}$. Training was conducted on 16 NVIDIA H20 GPUs for 4 hours. Detailed hyperparameters are provided in the Appendix \ref{apd:sft}.

\subsection{Response Optimization}
Inspired by recent work in rule-based reinforcement learning~\cite{guo2025deepseek}, we further optimized $U_{bc}$ using the Proximal Policy Optimization (PPO) algorithm~\cite{schulman2017proximal} to improve adherence to the requirements of vagueness and the format constraints. We defined a hybrid reward function as follows:
\begin{itemize}
\item $r_{format}$: A positive reward is granted if the model output  strictly follows the required JSON schema.
\item $r_{correctness}$: If the format is valid, we compare the generated fields (\texttt{action}, \texttt{name}, \texttt{return}) with the ground truth. 
\end{itemize}


We utilized the veRL framework~\cite{sheng2024hybridflow} for PPO training. The training was conducted on a cluster of 8 H20 GPUs. The PPO optimization process required approximately 8 hours to complete. Specific RL hyperparameters are listed in the Appendix \ref{apd:ppo}.



\subsection{User Simulator Evaluation} 
To investigate the performance differences between reinforcement learning strategies, we additionally implemented the Group Relative Policy Optimization (GRPO) algorithm~\cite{shao2024deepseekmath} with a group size of $N=4$.

\begin{table*}[h]
    \centering
    \small 
    \setlength{\tabcolsep}{3pt}
    \begin{tabular}{lccccccccc}
        \toprule
        \multirow{2}{*}{\textbf{Model}} & \multicolumn{4}{c}{\textbf{Error Action(\%)}} & \multicolumn{4}{c}{\textbf{Correct(\%)}} & \multirow{2}{*}{\textbf{Avg Turns}} \\
        \cmidrule(lr){2-5} \cmidrule(lr){6-9}
         & Path & Plan & Solution & Graph & Level 1 & Level 2 & Level 3 & Total & \\
        \midrule
        DeepSeek-V3.2      & 11.62 & 57.94 & 14.75 & 15.70 & 55.74 & 37.90 & 30.95 & 41.40 & 52.04 \\
        Qwen3-32b          & 11.82 & 36.50 & 44.21 & 7.47  & 34.43 & 28.23 & 24.60 & 29.03 & 43.17 \\
        Claude Sonnet 4.5  & 14.64 & 60.49 & 14.18 & 10.69 & 65.57 & 50.00 & \textbf{43.65} & 52.96 & 50.85 \\
        Gemini 2.5 pro     & 9.95  & 48.57 & 28.19 & 13.30 & \textbf{70.49} & 50.00 & 41.27 & \textbf{53.76} & 52.10 \\
        GPT-5         & 15.02 & 54.13 & 17.74 & 13.11 & 56.56 & \textbf{50.81} & \textbf{43.65} & 50.27 & 50.03 \\
        \bottomrule
    \end{tabular}
    \caption{Performance Summary of Different Models on JFTA-Bench (Subset)}
    \label{tab:main_results}
\end{table*}

We evaluate the trained user simulator from both functional correctness and behavioral realism perspectives. 
Correctness is measured in a strict, deterministic manner: a generated response is considered correct only when its \texttt{action}, \texttt{name}, and \texttt{return} fields exactly match the ground-truth annotation.

Beyond correctness, we assess the realism of generated user responses using an LLM-as-a-Judge protocol. 
The judge evaluates each response along two dimensions: \emph{fuzziness compliance}, which measures whether the response follows an indirect and non-diagnostic observation style, and \emph{naturalness}, which measures whether the expression resembles spontaneous human speech rather than templated output. 
We use GPT-4o as the judge model, with scores ranging from 0 to 3 for fuzziness and from 0 to 2 for naturalness. 
The detailed judging prompt is provided in Appendix~\ref{apd:user-eval}, and an additional human--LLM judge consistency analysis is reported in Appendix~\ref{apd:judge-consistency}.


\begin{table}[t]
\centering
\begin{tabular}{lccc}
\toprule
Model & Fuzziness & Naturalness & Correctness \\
\midrule
Base & 0.63 & 0.94 & 91.97 \\
SFT  & 2.24 & 1.51 & 99.94 \\
PPO  & \textbf{2.29} & \textbf{1.53} & \textbf{99.98} \\
GRPO & 2.21 & 1.46 & 99.47 \\
\bottomrule
\end{tabular}
\caption{Evaluation of user response quality. Base model refers to Qwen3-8B}
\label{tab:user_eval}
\end{table}


As shown in Table~\ref{tab:user_eval}, the base model performs poorly in both fuzziness and naturalness while All fine-tuned models achieve substantial improvements. SFT already produces more realistic user responses but tends to remain slightly rigid. PPO yields the best overall performance, whereas GRPO, despite comparable correctness, shows a mild regression in behavioral realism. Notably, although our reward design explicitly focuses only on format compliance and factual correctness, both fuzziness and naturalness improve after reinforcement learning. We hypothesize that reinforcement learning reshapes the response search space, mitigating the rigid response patterns induced by SFT and enabling more flexible, human-like expression. 


\section{Experiments}

\subsection{Main Evaluation}

We benchmarked a selection of frontier Large Language Models on JFTA-Bench. To optimize computational efficiency, initial experiments were conducted on a representative subset of the data,and the analysis of subset consistency is provided in Section \ref{sec:subset}. We evaluate a diverse set of state-of-the-art LLMs, comprising proprietary models, including Gemini 2.5 pro, GPT-5, and Claude Sonnet 4.5, alongside leading open-source counterparts including Qwen3-32B~\cite{yang2025qwen3} and DeepSeek-V3.2.

Table \ref{tab:main_results} summarizes the performance of different models in the JFTA-Bench. Among all evaluated models, Gemini 2.5 pro achieves the best overall performance, obtaining the highest correctness rate. In general, closed-source models consistently outperform open-source models across. Within the open-source group, DeepSeek-V3.2 demonstrates the strongest performance, achieving the highest correctness and more balanced error distribution.


\subsection{Analysis}
\label{sec:analyze}
As shown in Table \ref{tab:main_results}, the primary failure mode of the assistant lies in \textbf{Planning} errors(counted at a node level). Specifically, the assistant may repeatedly ask about a node for which the user has already provided diagnostic feedback. Notably, the think field of model's response often indicates that the model is already capable of determining whether a fault exists, yet it still fails to advance the dialogue. Furthermore, in almost all instances where a path error occurs, the model fails to recognize the mistake and make a correction. 
In addition, model shows insufficient discrimination between user feedback provided before and after the rollback, leading to unsuccessful error recovery. These behaviors suggest that the model may not sufficiently exploit dialogue-based memory when making decisions, potentially due to limitations in contextual attention over long interaction histories.

A second category of failures involves \textbf{Path} deviation, including but not limited to prematurely exiting a subtree before completing its diagnosis, fails to accurately identify the specific node that the user intends to revise within the dialogue history during rollback. To further quantify the impact of rollback scenarios, we conduct an ablation study in Appendix~\ref{apd:rollback-ablation}.

In a small number of cases, misinterpreting the \textbf{Graph} structure of the fault tree itself. These issues are likely caused by the model's limited understanding of highly complex fault tree structures, insufficient grasp of DFS traversal logic, and occasional misjudgments when reasoning over different logical gates.

Additionally, some failures arise at the \textbf{Solution} stage, where the assistant either provides an incorrect resolution for the remaining fault or outputs a solution associated with an incorrect identifier. 

Representative failure cases corresponding to the above error types are provided in Appendix~\ref{apd:failure-cases}.

\subsection{Ablation Studies}


\begin{figure}[htbp]
    \centering
    \includegraphics[width=0.9\linewidth]{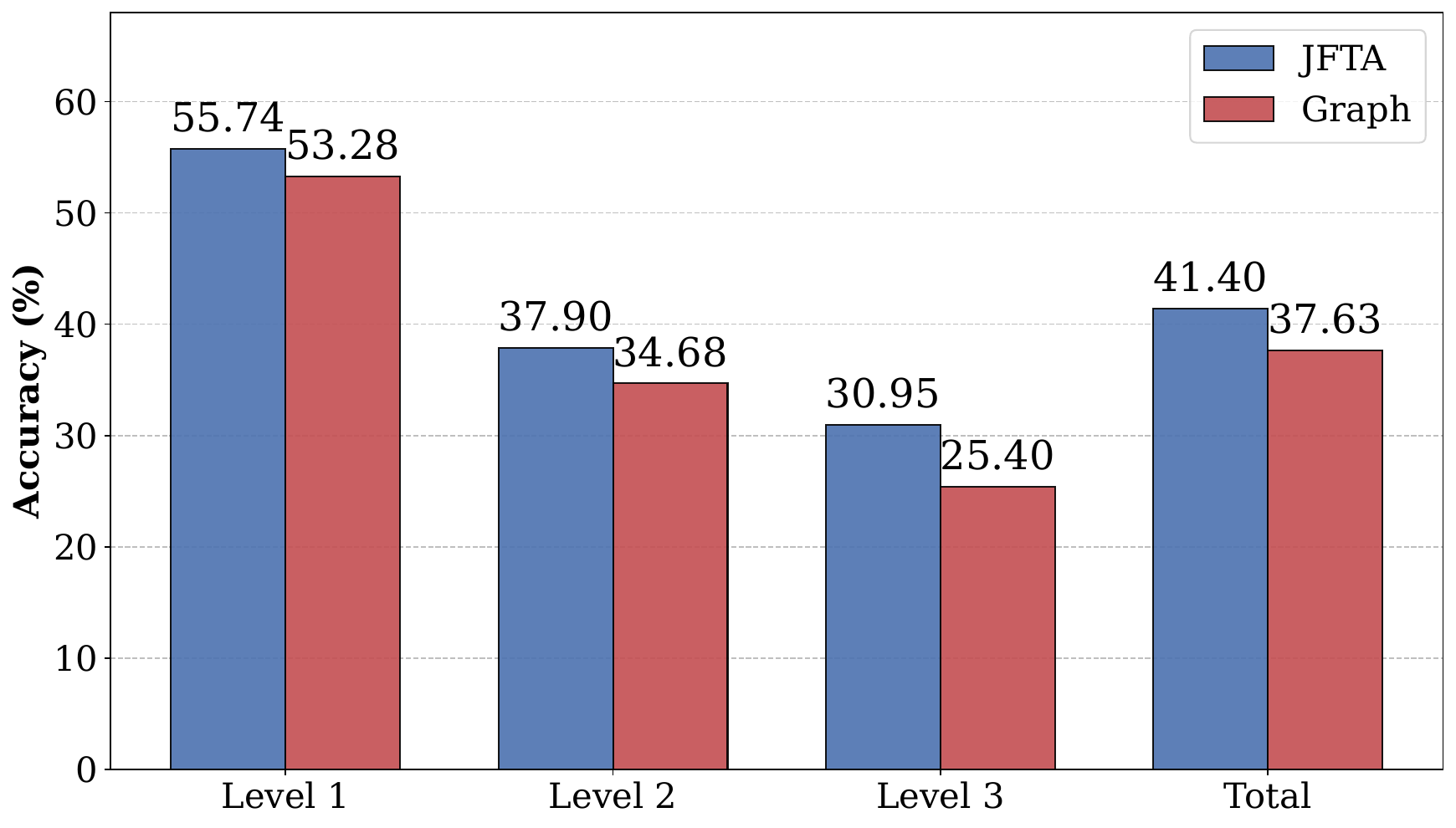}
    \caption{Accuracy comparison between JFTA and Node-Edge representations across different difficulty levels and overall performance.}
    \label{fig:ablation_rep}
\end{figure}

\paragraph{Effectiveness of JFTA Representation.} To demonstrate the superiority of our proposed syntax, we compared JFTA against a standard baseline representation that encodes the fault tree structure using a list of nodes and edges (Graph format). We evaluated the performance of DeepSeek-V3.2 on JFTA-Bench using both formats.  The Node-Edge representation leads to substantially longer prompts due to its verbosity, with an average length of 14,669.33 tokens under the Qwen3-8B tokenizer, compared to only 5,932.7 tokens for JFTA. As illustrated in Figure \ref{fig:ablation_rep}, across all difficulty levels, JFTA consistently achieves higher accuracy than the Node-Edge representation, and the performance gap further widens as task difficulty increases. This substantial reduction in token length allows more of the effective context window to be devoted to semantically meaningful structural information rather than redundant relational descriptors, thereby improving the model's ability to preserve long-range dependencies and perform multi-step reasoning. 

We additionally conduct a small-scale comparison against purely image-based fault tree inputs, detailed in Appendix~\ref{apd:image-ablation}. 
Even under a simplified path verification setting, image-based inputs fail to recover correct diagnostic paths, while JFTA achieves near-perfect performance.

\begin{table}[htbp]
  \centering
  \begin{tabular}{lcc}
    \toprule
    \textbf{Dataset} & \textbf{Average Turns} & \textbf{Correct (\%)} \\
    \midrule
    Subset 1  & 52.04 & 41.40 \\
    Subset 2  & 53.15 & 41.13 \\
    Subset 3  & 51.76 & 41.94 \\
    \midrule
    \textbf{Full Dataset} & \textbf{54.85} & \textbf{40.54} \\
    \bottomrule
  \end{tabular}
  \caption{Performance comparison across different subsets and the full dataset.}
  \label{tab:subset_consistency}
\end{table}

\paragraph{Subset Consistency Analysis.} 
\label{sec:subset}
Conducting full-scale evaluations on the entire JFTA-Bench dataset incurs substantial computational overhead. To mitigate this without compromising evaluation reliability, we constructed a representative test subset. Specifically, we randomly sampled entries from each level for all fault trees to form a subset containing 372 test entries. To verify the statistical consistency of this approach, we compared the performance of DeepSeek-V3.2 across the full dataset and three independently sampled subsets. The results, presented in Table \ref{tab:subset_consistency}, demonstrate that the absolute performance variance between different subsets remains less than 0.54\%, and the deviation between any subset and the full dataset is less than 1.40\%. This confirms that our subset evaluation strategy provides a robust and reliable approximation of the model's overall capability.

\begin{figure}[htbp]
    \centering
    \includegraphics[width=0.9\linewidth]{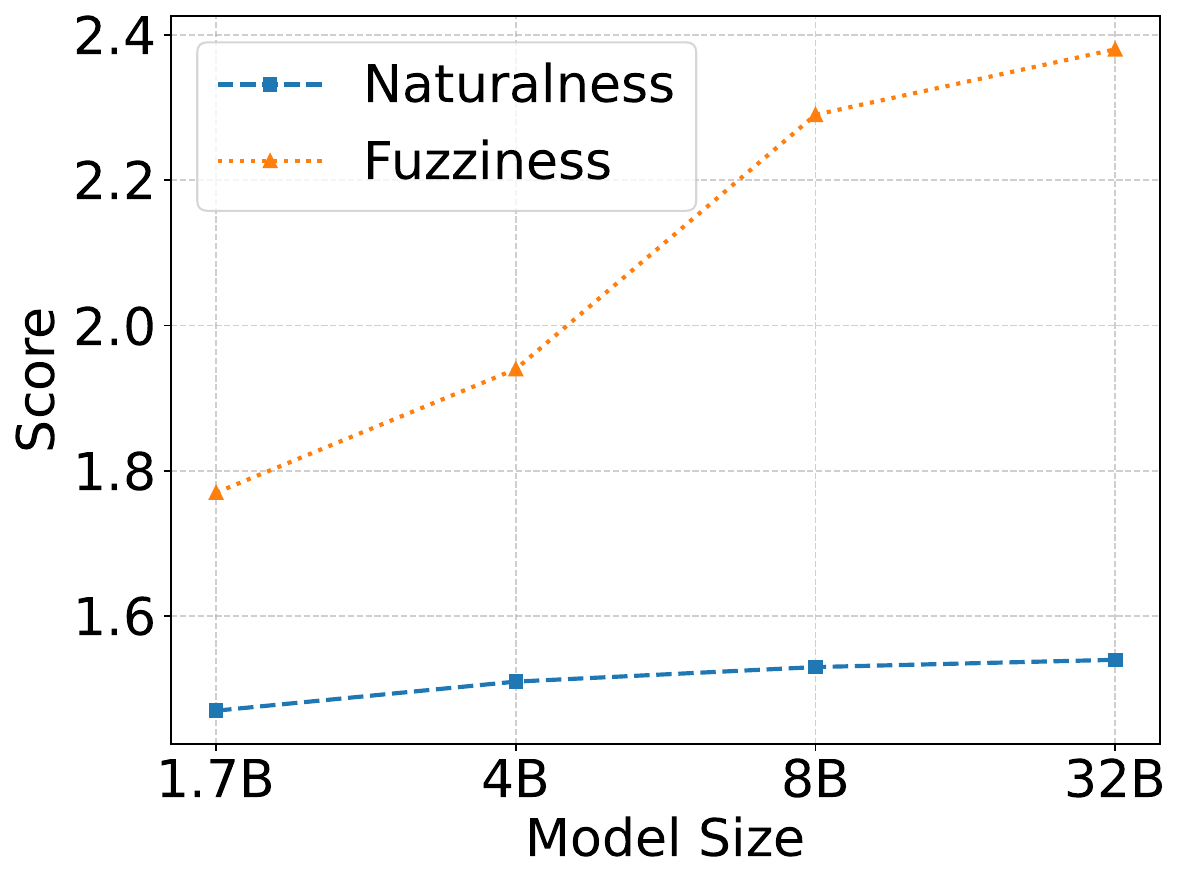}
    \caption{Performance of naturalness and fuzziness across different model sizes.}
    \label{fig:user_size}
\end{figure}

\paragraph{Impact of the Size of User Simulator.} We investigated the impact of model size scaling on the User Simulator's performance by comparing base models of varying sizes. We ultimately selected Qwen3-8B as the backbone for our user simulator. As illustrated in Figure \ref{fig:user_size}, after Behavior Cloning and Response Optimization, the 8B model significantly outperforms smaller variants in terms of response information quality. Conversely, while larger models offer marginal performance gains, they incur substantially higher training and inference costs. The 8B parameter count proves to be the optimal balance, delivering robust simulation capabilities with high computational efficiency.

\section{Conclusion}

In this work, we bridge the gap between visual fault tree analysis and large language models by introducing JFTA, a structured textual representation that supports rigorous logical reasoning. Building on this syntax, we construct JFTA-Bench, a comprehensive multi-turn dialogue benchmark that evaluates model performance in complex diagnostic scenarios, featuring a novel evaluation of error recovery capabilities via long-range state rollback, and a user simulator trained for generating diverse and vague responses. The benchmark highlights the remaining challenges for LLMs in maintaining consistency during dynamic, error-prone interactions and planning ability. This work lays a solid foundation for future research on LLMs in real-world applications, while also providing new insights into the analysis and practical use of LLMs in JFTA-based malfunction diagnosis and reasoning.

\section*{Limitations}

Although the proposed JFTA and its benchmark perform well in evaluations, several limitations remain. First, due to strict confidentiality protocols in the industrial sector, our benchmark lacks large-scale, native fault tree data collected directly from real-world production lines. Consequently, the current semi-synthetic dataset may not fully capture the unpredictable noise and complexity of rare industrial malfunctions, which restricts a comprehensive evaluation of a model's diagnostic reasoning capabilities in real-world scenarios. Second, the long-horizon, multi-turn interactions required for evaluation incur substantial computational overhead, necessitating the use of smaller representative subsets for certain analyses. Finally, JFTA has not yet been deployed within operational industrial maintenance software; its practical integration challenges, real-time performance, and usability in live production environments remain to be validated. Our dataset is used solely for academic research.

\section*{Acknowledgments}

The authors wish to thank the anonymous reviewers for their helpful comments. This work was partially funded by National Key R\&D Program of China No.2025ZD0215702, National Natural Science Foundation of China (No. 62576106, 62476061, 62376061).

\bibliography{custom}

@inproceedings{singh2025openai,
  title={OpenAI GPT-5 System Card},
  author={Aaditya K. Singh and Adam Fry and Adam Perelman and Adam Tart and Adithya Ganesh and Ahmed El-Kishky and Aidan McLaughlin and Aiden Low and AJ Ostrow and Akhila Ananthram and Akshay Nathan and Alan Luo and Alec Helyar and Aleksander Madry and Aleksandr A Efremov and Aleksandra Spyra and Alex Baker-Whitcomb and Alex Beutel and Alex Karpenko and Aleksandar Makelov and Alexander Neitz and Alexander Z. Wei and Alexandra Barr and Alexandre Kirchmeyer and A.A. Ivanov and Alexi Christakis and A. Gillespie and Allison Tam and Ally Bennett and Alvin Wan and Alyssa Huang and Amy McDonald Sandjideh and Amy Yang and Ananya Kumar and Andre Saraiva and Andrea Vallone and A.S. Gheorghe and Andr{\'e}s G. Garcia and Andrew Braunstein and Andrew Liu and Andrew M. Schmidt and Andrey Mereskin and Andrey Mishchenko and Andy Applebaum and Andrew Rogerson and Abhirami Rajan and Annie Y. Wei and Anoop Kotha and Anubhav Srivastava and A. Agrawal and Arun Vijayvergiya and Ashley Tyra and Ashvin Nair and Avi Nayak and Ben Eggers and Bessie Ji and Beth Hoover and Bill Chen and Blair Chen and Boaz Barak and Borys Minaiev and Botao Hao and Bowen Baker and Brad Lightcap and Brandon McKinzie and Brandon Wang and Brendan Quinn and Brian Fioca and Brian Hsu and Brian Yang and Brian Yu and Brian Zhang and B. Brenner and Callie Riggins Zetino and Cameron Raymond and Camillo Lugaresi and Carolina Paz and Cary Hudson and Cedric Whitney and Chak Li and Charles Chen and Charlotte Cole and Chelsea Voss and Chen Ding and Chen Shen and Chengdu Huang and Chris Colby and Chris Hallacy and Chris Koch and Chris Lu and Chris Kaplan and Christina Kim and CJ Minott-Henriques and Cliff Frey and Cody W. Yu and Coley Czarnecki and Colin Reid and Colin Wei and Cory Decareaux and Cristina Scheau and Cyril Zhang and Cyrus Forbes and Da Tang and D.P Goldberg and Dan Roberts and Dana Palmie and Daniel Kappler and Daniel Levine and Daniel Wright and David Leo and David Lin and David Robinson and Declan Grabb and Derek Chen and Derek Lim and Derek Salama and Dibyatanoy Bhattacharjee and Dimitris Tsipras and Dinghua Li and Dingli Yu and DJ Strouse and D.R. Williams and Dylan Hunn and Edward Bayes and Edwin Arbus and Ekin Aky{\"u}rek and Elaine Ya Le and Elana Widmann and Eli Yani and Elizabeth Proehl and Enis Sert and Enoch Cheung and Eric M. Schwartz and Eric Han and Eric Jiang and Eric Mitchell and Eric Sigler and Eric Wallace and Erik Ritter and Erin Kavanaugh and Evan Mays and Evgenii Nikishin and Fangyuan Li and Felipe Petroski Such and Filipe de Avila Belbute Peres and Filippo Raso and Florent Bekerman and Foivos Tsimpourlas and Fotis Chantzis and Francis Song and Francis Zhang and Gaby Raila and Garrett McGrath and Gary J. Briggs and Gary Yang and Giambattista Parascandolo and Gildas Chabot and Grace Kim and Grace Zhao and Gregory Valiant and Guillaume Leclerc and Hadi Salman and Hanson Wang and Hao Sheng and Hao-Yang Jiang and Haoyu Wang and Haozhun Jin and Harshit S. Sikchi and Heather Schmidt and Henry Aspegren and Honglin Chen and Huida Qiu and Hunter Lightman and Ian Connick Covert and Ian Kivlichan and Ian Silber and Ian Sohl and Ibrahim Hammoud and Ignasi Clavera and Ikai Lan and Ilge Akkaya and Ilya Kostrikov and Irina Kofman and Isak Czeresnia Etinger and Ishaan Singal and Jackie Hehir and Jacob Huh and Jacqueline Pan and J. Wilczynski and Jakub W. Pachocki and James Lee and J. Quinn and Jamie Ryan Kiros and Janvi Kalra and Jasmyn Samaroo and Jason Wang and Jason Wolfe and Jay Chen and Jason Wang and Jean Harb and Jeff Han and Jeffrey Wang and Jennifer Zhao and Jeremy Chen and Jerene Zhe Yang and Jerry Tworek and Jesse Chand and Jessica Landon and Jessica Y. Liang and Ji Lin and Jiancheng Liu and Jianfeng Wang and Jie Tang and Jihan Yin and Joanne Jang and Joel M. Morris and Joey Flynn and Johannes Ferstad and Johannes Heidecke and John Fishbein and John Hallman and Jon E Grant and Jonathan Chien and Jonathan Gordon and Jongsoo Park and Jordan Liss and Jos Kraaijeveld and Joseph Guay and Joseph Mo and Joshua Lawson and Josh McGrath and Joshua Vendrow and Joy Jiao and Julian Hui Min Lee and Julie Steele and Juliette Wang and Junhua Mao and Kai Chen and Kai Hayashi and Kai Xiao and Kamyar Salahi and Kan Wu and Karan Sekhri and Karan Sharma and Karan Singhal and Karen Li and Kenny Nguyen and Keren Gu-Lemberg and Kevin King and Kevin Liu and Kevin R. Stone and Kevin Yu and Kristen Ying and Kristian Georgiev and Kelvin Lim and Kushal Tirumala and Kyle Miller and Lama Ahmad and Larry Lv and L. Clare and Laurance Fauconnet and Lauren Itow and Lauren Yang and Laurentia M. Romaniuk and Leah Anise and Lee Byron and Leher Pathak and Leon Maksin and Leyan Lo and Leyton Ho and Li Jing and Liang Wu and Lianggeng Xiong and Lien Mamitsuka and Lin Yang and Lindsay McCallum and Lindsey Held and llann Bourgeois and Logan Engstrom and Lorenz Kuhn and Louis Feuvrier and Lu Zhang and Lucas Switzer and Lukasz Kondraciuk and Lukasz Kaiser and Manas R. Joglekar and Mandeep Singh and M Shah and Manuka Stratta and Marcus Williams and Mark Chen and Mark Sun and Marselus Cayton and Martin Li and Marvin Zhang and Marwan Aljubeh and Matt Nichols and Matthew Haines and Max Schwarzer and Mayank Mike Gupta and Meghan Shah and Melody Y. Huang and Meng Dong and Mengqing Wang and Mia Glaese and Micah Carroll and Michael Lampe and Michael Malek and Michael Sharman and Michael Zhang and Michele Wang and Michelle Pokrass and Mihai Florian and Mikhail Pavlov and Miles Wang and Ming Feng Chen and Mingxuan Wang and Min Feng and Mo Bavarian and Molly Lin and Moose Abdool and Mostafa Rohaninejad and Nacho Soto and Natalie M. Staudacher and Natan LaFontaine and Nathan Marwell and Nelson F. Liu and Nick Preston and Nick Turley and Nicklas Ansman and Nicole Blades and Nikil Pancha and Nikita Mikhaylin and Niko Felix and Nikunj Handa and Nishant Rai and Nitish Shirish Keskar and Noam Brown and Ofir Nachum and Oleg Boiko and Oleg Murk and Olivia Watkins and Oona Gleeson and Pamela Mishkin and Patryk Lesiewicz and Paul Baltescu and Pavel Belov and Peter Zhokhov and Philip Pronin and Phillip Guo and Phoebe Thacker and Qi Liu and Qim-ing Yuan and Qinghua Liu and Rachel Dias and Rachel Puckett and Rahul K. Arora and Ravi Teja Mullapudi and Raz Gaon and Reah Miyara and Ren Song and Rishabh Aggarwal and RJ Marsan and Robel Yemiru and Robert Xiong and Rohan Kshirsagar and Rohan Nuttall and Roman Tsiupa and Ronen Eldan and Rose Wang and Roshan James and R. Ziv and Rui Shu and Ruslan T. Nigmatullin and Saachi Jain and Saam Talaie and Sam Altman and Sam Arnesen and Sam Toizer and Sam Toyer and Samuel Miserendino and Sandhini Agarwal and Sarah Yoo and Savannah Heon and Scott Ethersmith and Sean Grove and Sean Taylor and S{\'e}bastien Bubeck and Sever Banesiu and Shaokyi Amdo and Shengjia Zhao and Sherwin Wu and Shibani Santurkar and Shiyu Zhao and Shraman Ray Chaudhuri and Shreyas Krishnaswamy and Shuaiqi Xia and Shuyan Cheng and Shyamal Anadkat and Sim'on Posada Fishman and Simon Tobin and Siyuan Fu and Somay Jain and Song Mei and Sonya Egoian and Spencer Kim and Simone L. van Golden and SQ Mah and Stephanie L. Lin and Stephen Imm and Steve Sharpe and Steve Yadlowsky and Sulman Choudhry and Su-hyeon Eum and Suvansh Sanjeev and Tabarak Khan and Tal Stramer and Tao Wang and Tao Xin and Tarun Gogineni and Taya Christianson and Ted Sanders and Tejal Patwardhan and Thomas Degry and Thomas Shadwell and Tianfu Fu and Tianshi Gao and T. Garipov and Tina Sriskandarajah and Toki Sherbakov and Tomer Kaftan and Tomoya Hiratsuka and Tongzhou Wang and Tony Song and Tony Zhao and Troy A. Peterson and V. V. Kharitonov and Victoria Chernova and Vineet Kosaraju and Vishal Kuo and Vitchyr H. Pong and Vivek Verma and Vladimir G. Petrov and Wanning Jiang and Weixin Zhang and Wenda Zhou and Wen-Li Xie and Wenting Zhan and Wes McCabe and Will DePue and Will Ellsworth and Wulfie Bain and Wyatt Thompson and Xiangning Chen and Xi-Meng Qi and Xin Xiang and Xinwei Shi and Yann Dubois and Yaodong Yu and Yara Khakbaz and Yifan Wu and Yilei Qian and Yin Tat Lee and Yinbo Chen and Yizhen Zhang and Yi-Qu Xiong and Yonglong Tian and Young Shin Cha and Yu Bai and Yu Yang and Yuan Yuan and Yuanzhi Li and Yufeng Zhang and Yuguang Yang and Yujia Jin and Yun Jiang and Yunyun Wang and Yushi Wang and Yutian Liu and Zach Stubenvoll and Zehao Dou and Zheng Wu and Zhigang Wang},
  year={2025},
  url={https://api.semanticscholar.org/CorpusID:284532660}
}

@misc{comanici2025gemini25pushingfrontier,
      title={Gemini 2.5: Pushing the Frontier with Advanced Reasoning, Multimodality, Long Context, and Next Generation Agentic Capabilities}, 
      author={Gheorghe Comanici and Eric Bieber and Mike Schaekermann and Ice Pasupat and Noveen Sachdeva and Inderjit Dhillon and Marcel Blistein and Ori Ram and Dan Zhang and Evan Rosen and Luke Marris and Sam Petulla and Colin Gaffney and Asaf Aharoni and Nathan Lintz and Tiago Cardal Pais and Henrik Jacobsson and Idan Szpektor and Nan-Jiang Jiang and Krishna Haridasan and Ahmed Omran and Nikunj Saunshi and Dara Bahri and Gaurav Mishra and Eric Chu and Toby Boyd and Brad Hekman and Aaron Parisi and Chaoyi Zhang and Kornraphop Kawintiranon and Tania Bedrax-Weiss and Oliver Wang and Ya Xu and Ollie Purkiss and Uri Mendlovic and Ilaï Deutel and Nam Nguyen and Adam Langley and Flip Korn and Lucia Rossazza and Alexandre Ramé and Sagar Waghmare and Helen Miller and Nathan Byrd and Ashrith Sheshan and Raia Hadsell and Sangnie Bhardwaj and Pawel Janus and Tero Rissa and Dan Horgan and Alvin Abdagic and Lior Belenki and James Allingham and Anima Singh and Theo Guidroz and Srivatsan Srinivasan and Herman Schmit and Kristen Chiafullo and Andre Elisseeff and Nilpa Jha and Prateek Kolhar and Leonard Berrada and Frank Ding and Xiance Si and Shrestha Basu Mallick and Franz Och and Sofia Erell and Eric Ni and Tejasi Latkar and Sherry Yang and Petar Sirkovic and Ziqiang Feng and Robert Leland and Rachel Hornung and Gang Wu and Charles Blundell and Hamidreza Alvari and Po-Sen Huang and Cathy Yip and Sanja Deur and Li Liu and Gabriela Surita and Pablo Duque and Dima Damen and Johnson Jia and Arthur Guez and Markus Mircea and Animesh Sinha and Alberto Magni and Paweł Stradomski and Tal Marian and Vlado Galić and Wenhu Chen and Hisham Husain and Achintya Singhal and Dominik Grewe and François-Xavier Aubet and Shuang Song and Lorenzo Blanco and Leland Rechis and Lewis Ho and Rich Munoz and Kelvin Zheng and Jessica Hamrick and Kevin Mather and Hagai Taitelbaum and Eliza Rutherford and Yun Lei and Kuangyuan Chen and Anand Shukla and Erica Moreira and Eric Doi and Berivan Isik and Nir Shabat and Dominika Rogozińska and Kashyap Kolipaka and Jason Chang and Eugen Vušak and Srinivasan Venkatachary and Shadi Noghabi and Tarun Bharti and Younghoon Jun and Aleksandr Zaks and Simon Green and Jeshwanth Challagundla and William Wong and Muqthar Mohammad and Dean Hirsch and Yong Cheng and Iftekhar Naim and Lev Proleev and Damien Vincent and Aayush Singh and Maxim Krikun and Dilip Krishnan and Zoubin Ghahramani and Aviel Atias and Rajeev Aggarwal and Christo Kirov and Dimitrios Vytiniotis and Christy Koh and Alexandra Chronopoulou and Pawan Dogra and Vlad-Doru Ion and Gladys Tyen and Jason Lee and Felix Weissenberger and Trevor Strohman and Ashwin Balakrishna and Jack Rae and Marko Velic and Raoul de Liedekerke and Oded Elyada and Wentao Yuan and Canoee Liu and Lior Shani and Sergey Kishchenko and Bea Alessio and Yandong Li and Richard Song and Sam Kwei and Orion Jankowski and Aneesh Pappu and Youhei Namiki and Yenai Ma and Nilesh Tripuraneni and Colin Cherry and Marissa Ikonomidis and Yu-Cheng Ling and Colin Ji and Beka Westberg and Auriel Wright and Da Yu and David Parkinson and Swaroop Ramaswamy and Jerome Connor and Soheil Hassas Yeganeh and Snchit Grover and George Kenwright and Lubo Litchev and Chris Apps and Alex Tomala and Felix Halim and Alex Castro-Ros and Zefei Li and Anudhyan Boral and Pauline Sho and Michal Yarom and Eric Malmi and David Klinghoffer and Rebecca Lin and Alan Ansell and Pradeep Kumar S and Shubin Zhao and Siqi Zuo and Adam Santoro and Heng-Tze Cheng and Solomon Demmessie and Yuchi Liu and Nicole Brichtova and Allie Culp and Nathaniel Braun and Dan Graur and Will Ng and Nikhil Mehta and Aaron Phillips and Patrik Sundberg and Varun Godbole and Fangyu Liu and Yash Katariya and David Rim and Mojtaba Seyedhosseini and Sean Ammirati and Jonas Valfridsson and Mahan Malihi and Timothy Knight and Andeep Toor and Thomas Lampe and Abe Ittycheriah and Lewis Chiang and Chak Yeung and Alexandre Fréchette and Jinmeng Rao and Huisheng Wang and Himanshu Srivastava and Richard Zhang and Rocky Rhodes and Ariel Brand and Dean Weesner and Ilya Figotin and Felix Gimeno and Rachana Fellinger and Pierre Marcenac and José Leal and Eyal Marcus and Victor Cotruta and Rodrigo Cabrera and Sheryl Luo and Dan Garrette and Vera Axelrod and Sorin Baltateanu and David Barker and Dongkai Chen and Horia Toma and Ben Ingram and Jason Riesa and Chinmay Kulkarni and Yujing Zhang and Hongbin Liu and Chao Wang and Martin Polacek and Will Wu and Kai Hui and Adrian N Reyes and Yi Su and Megan Barnes and Ishaan Malhi and Anfal Siddiqui and Qixuan Feng and Mihai Damaschin and Daniele Pighin and Andreas Steiner and Samuel Yang and Ramya Sree Boppana and Simeon Ivanov and Arun Kandoor and Aditya Shah and Asier Mujika and Da Huang and Christopher A. Choquette-Choo and Mohak Patel and Tianhe Yu and Toni Creswell and Jerry and Liu and Catarina Barros and Yasaman Razeghi and Aurko Roy and Phil Culliton and Binbin Xiong and Jiaqi Pan and Thomas Strohmann and Tolly Powell and Babi Seal and Doug DeCarlo and Pranav Shyam and Kaan Katircioglu and Xuezhi Wang and Cassidy Hardin and Immanuel Odisho and Josef Broder and Oscar Chang and Arun Nair and Artem Shtefan and Maura O'Brien and Manu Agarwal and Sahitya Potluri and Siddharth Goyal and Amit Jhindal and Saksham Thakur and Yury Stuken and James Lyon and Kristina Toutanova and Fangxiaoyu Feng and Austin Wu and Ben Horn and Alek Wang and Alex Cullum and Gabe Taubman and Disha Shrivastava and Chongyang Shi and Hamish Tomlinson and Roma Patel and Tao Tu and Ada Maksutaj Oflazer and Francesco Pongetti and Mingyao Yang and Adrien Ali Taïga and Vincent Perot and Nuo Wang Pierse and Feng Han and Yoel Drori and Iñaki Iturrate and Ayan Chakrabarti and Legg Yeung and Dave Dopson and Yi-ting Chen and Apoorv Kulshreshtha and Tongfei Guo and Philip Pham and Tal Schuster and Junquan Chen and Alex Polozov and Jinwei Xing and Huanjie Zhou and Praneeth Kacham and Doron Kukliansky and Antoine Miech and Sergey Yaroshenko and Ed Chi and Sholto Douglas and Hongliang Fei and Mathieu Blondel and Preethi Myla and Lior Madmoni and Xing Wu and Daniel Keysers and Kristian Kjems and Isabela Albuquerque and Lijun Yu and Joel D'sa and Michelle Plantan and Vlad Ionescu and Jaume Sanchez Elias and Abhirut Gupta and Manish Reddy Vuyyuru and Fred Alcober and Tong Zhou and Kaiyang Ji and Florian Hartmann and Subha Puttagunta and Hugo Song and Ehsan Amid and Anca Stefanoiu and Andrew Lee and Paul Pucciarelli and Emma Wang and Amit Raul and Slav Petrov and Isaac Tian and Valentin Anklin and Nana Nti and Victor Gomes and Max Schumacher and Grace Vesom and Alex Panagopoulos and Konstantinos Bousmalis and Daniel Andor and Josh Jacob and Yuan Zhang and Bill Rosgen and Matija Kecman and Matthew Tung and Alexandra Belias and Noah Goodman and Paul Covington and Brian Wieder and Nikita Saxena and Elnaz Davoodi and Muhuan Huang and Sharath Maddineni and Vincent Roulet and Folawiyo Campbell-Ajala and Pier Giuseppe Sessa and Xintian and Wu and Guangda Lai and Paul Collins and Alex Haig and Vytenis Sakenas and Xiaowei Xu and Marissa Giustina and Laurent El Shafey and Pichi Charoenpanit and Shefali Garg and Joshua Ainslie and Boone Severson and Montse Gonzalez Arenas and Shreya Pathak and Sujee Rajayogam and Jie Feng and Michiel Bakker and Sheng Li and Nevan Wichers and Jamie Rogers and Xinyang Geng and Yeqing Li and Rolf Jagerman and Chao Jia and Nadav Olmert and David Sharon and Matthew Mauger and Sandeep Mariserla and Hongxu Ma and Megha Mohabey and Kyuyeun Kim and Alek Andreev and Scott Pollom and Juliette Love and Vihan Jain and Priyanka Agrawal and Yannick Schroecker and Alisa Fortin and Manfred Warmuth and Ji Liu and Andrew Leach and Irina Blok and Ganesh Poomal Girirajan and Roee Aharoni and Benigno Uria and Andrei Sozanschi and Dan Goldberg and Lucian Ionita and Marco Tulio Ribeiro and Martin Zlocha and Vighnesh Birodkar and Sami Lachgar and Liangzhe Yuan and Himadri Choudhury and Matt Ginsberg and Fei Zheng and Gregory Dibb and Emily Graves and Swachhand Lokhande and Gabriel Rasskin and George-Cristian Muraru and Corbin Quick and Sandeep Tata and Pierre Sermanet and Aditya Chawla and Itay Karo and Yan Wang and Susan Zhang and Orgad Keller and Anca Dragan and Guolong Su and Ian Chou and Xi Liu and Yiqing Tao and Shruthi Prabhakara and Marc Wilson and Ruibo Liu and Shibo Wang and Georgie Evans and David Du and Alfonso Castaño and Gautam Prasad and Mona El Mahdy and Sebastian Gerlach and Machel Reid and Jarrod Kahn and Amir Zait and Thanumalayan Sankaranarayana Pillai and Thatcher Ulrich and Guanyu Wang and Jan Wassenberg and Efrat Farkash and Kiran Yalasangi and Congchao Wang and Maria Bauza and Simon Bucher and Ting Liu and Jun Yan and Gary Leung and Vikas Sindhwani and Parker Barnes and Avi Singh and Ivan Jurin and Jichuan Chang and Niket Kumar Bhumihar and Sivan Eiger and Gui Citovsky and Ben Withbroe and Zhang Li and Siyang Xue and Niccolò Dal Santo and Georgi Stoyanov and Yves Raimond and Steven Zheng and Yilin Gao and Vít Listík and Sławek Kwasiborski and Rachel Saputro and Adnan Ozturel and Ganesh Mallya and Kushal Majmundar and Ross West and Paul Caron and Jinliang Wei and Lluis Castrejon and Sharad Vikram and Deepak Ramachandran and Nikhil Dhawan and Jiho Park and Sara Smoot and George van den Driessche and Yochai Blau and Chase Malik and Wei Liang and Roy Hirsch and Cicero Nogueira dos Santos and Eugene Weinstein and Aäron van den Oord and Sid Lall and Nicholas FitzGerald and Zixuan Jiang and Xuan Yang and Dale Webster and Ali Elqursh and Aedan Pope and Georges Rotival and David Raposo and Wanzheng Zhu and Jeff Dean and Sami Alabed and Dustin Tran and Arushi Gupta and Zach Gleicher and Jessica Austin and Edouard Rosseel and Megh Umekar and Dipanjan Das and Yinghao Sun and Kai Chen and Karolis Misiunas and Xiang Zhou and Yixian Di and Alyssa Loo and Josh Newlan and Bo Li and Vinay Ramasesh and Ying Xu and Alex Chen and Sudeep Gandhe and Radu Soricut and Nikita Gupta and Shuguang Hu and Seliem El-Sayed and Xavier Garcia and Idan Brusilovsky and Pu-Chin Chen and Andrew Bolt and Lu Huang and Alex Gurney and Zhiying Zhang and Alexander Pritzel and Jarek Wilkiewicz and Bryan Seybold and Bhargav Kanagal Shamanna and Felix Fischer and Josef Dean and Karan Gill and Ross Mcilroy and Abhishek Bhowmick and Jeremy Selier and Antoine Yang and Derek Cheng and Vladimir Magay and Jie Tan and Dhriti Varma and Christian Walder and Tomas Kocisky and Ryo Nakashima and Paul Natsev and Mike Kwong and Ionel Gog and Chiyuan Zhang and Sander Dieleman and Thomas Jimma and Andrey Ryabtsev and Siddhartha Brahma and David Steiner and Dayou Du and Ante Žužul and Mislav Žanić and Mukund Raghavachari and Willi Gierke and Zeyu Zheng and Dessie Petrova and Yann Dauphin and Yuchuan Liu and Ido Kessler and Steven Hand and Chris Duvarney and Seokhwan Kim and Hyo Lee and Léonard Hussenot and Jeffrey Hui and Josh Smith and Deepali Jain and Jiawei Xia and Gaurav Singh Tomar and Keyvan Amiri and Du Phan and Fabian Fuchs and Tobias Weyand and Nenad Tomasev and Alexandra Cordell and Xin Liu and Jonathan Mallinson and Pankaj Joshi and Andy Crawford and Arun Suggala and Steve Chien and Nick Fernando and Mariella Sanchez-Vargas and Duncan Williams and Phil Crone and Xiyang Luo and Igor Karpov and Jyn Shan and Terry Thurk and Robin Strudel and Paul Voigtlaender and Piyush Patil and Tim Dozat and Ali Khodaei and Sahil Singla and Piotr Ambroszczyk and Qiyin Wu and Yifan Chang and Brian Roark and Chaitra Hegde and Tianli Ding and Angelos Filos and Zhongru Wu and André Susano Pinto and Shuang Liu and Saarthak Khanna and Aditya Pandey and Siobhan Mcloughlin and Qiujia Li and Sam Haves and Allan Zhou and Elena Buchatskaya and Isabel Leal and Peter de Boursac and Nami Akazawa and Nina Anderson and Terry Chen and Krishna Somandepalli and Chen Liang and Sheela Goenka and Stephanie Winkler and Alexander Grushetsky and Yifan Ding and Jamie Smith and Fan Ye and Jordi Pont-Tuset and Eric Li and Ruichao Li and Tomer Golany and Dawid Wegner and Tao Jiang and Omer Barak and Yuan Shangguan and Eszter Vértes and Renee Wong and Jörg Bornschein and Alex Tudor and Michele Bevilacqua and Tom Schaul and Ankit Singh Rawat and Yang Zhao and Kyriakos Axiotis and Lei Meng and Cory McLean and Jonathan Lai and Jennifer Beattie and Nate Kushman and Yaxin Liu and Blair Kutzman and Fiona Lang and Jingchen Ye and Praneeth Netrapalli and Pushkar Mishra and Myriam Khan and Megha Goel and Rob Willoughby and David Tian and Honglei Zhuang and JD Chen and Zak Tsai and Tasos Kementsietsidis and Arjun Khare and James Keeling and Keyang Xu and Nathan Waters and Florent Altché and Ashok Popat and Bhavishya Mittal and David Saxton and Dalia El Badawy and Michael Mathieu and Zheng Zheng and Hao Zhou and Nishant Ranka and Richard Shin and Qingnan Duan and Tim Salimans and Ioana Mihailescu and Uri Shaham and Ming-Wei Chang and Yannis Assael and Nishanth Dikkala and Martin Izzard and Vincent Cohen-Addad and Cat Graves and Vlad Feinberg and Grace Chung and DJ Strouse and Danny Karmon and Sahand Sharifzadeh and Zoe Ashwood and Khiem Pham and Jon Blanton and Alex Vasiloff and Jarred Barber and Mark Geller and Aurick Zhou and Fedir Zubach and Tzu-Kuo Huang and Lei Zhang and Himanshu Gupta and Matt Young and Julia Proskurnia and Ronny Votel and Valentin Gabeur and Gabriel Barcik and Aditya Tripathi and Hongkun Yu and Geng Yan and Beer Changpinyo and Filip Pavetić and Amy Coyle and Yasuhisa Fujii and Jorge Gonzalez Mendez and Tianhao Zhou and Harish Rajamani and Blake Hechtman and Eddie Cao and Da-Cheng Juan and Yi-Xuan Tan and Valentin Dalibard and Yilun Du and Natalie Clay and Kaisheng Yao and Wenhao Jia and Dimple Vijaykumar and Yuxiang Zhou and Xinyi Bai and Wei-Chih Hung and Steven Pecht and Georgi Todorov and Nikhil Khadke and Pramod Gupta and Preethi Lahoti and Arnaud Autef and Karthik Duddu and James Lee-Thorp and Alexander Bykovsky and Tautvydas Misiunas and Sebastian Flennerhag and Santhosh Thangaraj and Jed McGiffin and Zack Nado and Markus Kunesch and Andreas Noever and Amir Hertz and Marco Liang and Victor Stone and Evan Palmer and Samira Daruki and Arijit Pramanik and Siim Põder and Austin Kyker and Mina Khan and Evgeny Sluzhaev and Marvin Ritter and Avraham Ruderman and Wenlei Zhou and Chirag Nagpal and Kiran Vodrahalli and George Necula and Paul Barham and Ellie Pavlick and Jay Hartford and Izhak Shafran and Long Zhao and Maciej Mikuła and Tom Eccles and Hidetoshi Shimokawa and Kanav Garg and Luke Vilnis and Hanwen Chen and Ilia Shumailov and Kuang-Huei Lee and Abdelrahman Abdelhamed and Meiyan Xie and Vered Cohen and Ester Hlavnova and Dan Malkin and Chawin Sitawarin and James Lottes and Pauline Coquinot and Tianli Yu and Sandeep Kumar and Jingwei Zhang and Aroma Mahendru and Zafarali Ahmed and James Martens and Tao Chen and Aviel Boag and Daiyi Peng and Coline Devin and Arseniy Klimovskiy and Mary Phuong and Danny Vainstein and Jin Xie and Bhuvana Ramabhadran and Nathan Howard and Xinxin Yu and Gitartha Goswami and Jingyu Cui and Sam Shleifer and Mario Pinto and Chih-Kuan Yeh and Ming-Hsuan Yang and Sara Javanmardi and Dan Ethier and Chace Lee and Jordi Orbay and Suyog Kotecha and Carla Bromberg and Pete Shaw and James Thornton and Adi Gerzi Rosenthal and Shane Gu and Matt Thomas and Ian Gemp and Aditya Ayyar and Asahi Ushio and Aarush Selvan and Joel Wee and Chenxi Liu and Maryam Majzoubi and Weiren Yu and Jake Abernethy and Tyler Liechty and Renke Pan and Hoang Nguyen and Qiong and Hu and Sarah Perrin and Abhinav Arora and Emily Pitler and Weiyi Wang and Kaushik Shivakumar and Flavien Prost and Ben Limonchik and Jing Wang and Yi Gao and Timothee Cour and Shyamal Buch and Huan Gui and Maria Ivanova and Philipp Neubeck and Kelvin Chan and Lucy Kim and Huizhong Chen and Naman Goyal and Da-Woon Chung and Lu Liu and Yao Su and Anastasia Petrushkina and Jiajun Shen and Armand Joulin and Yuanzhong Xu and Stein Xudong Lin and Yana Kulizhskaya and Ciprian Chelba and Shobha Vasudevan and Eli Collins and Vasilisa Bashlovkina and Tony Lu and Doug Fritz and Jongbin Park and Yanqi Zhou and Chen Su and Richard Tanburn and Mikhail Sushkov and Mitchelle Rasquinha and Jinning Li and Jennifer Prendki and Yiming Li and Pallavi LV and Shriya Sharma and Hen Fitoussi and Hui Huang and Andrew Dai and Phuong Dao and Mike Burrows and Henry Prior and Danfeng Qin and Golan Pundak and Lars Lowe Sjoesund and Art Khurshudov and Zhenkai Zhu and Albert Webson and Elizabeth Kemp and Tat Tan and Saurabh Agrawal and Susie Sargsyan and Liqun Cheng and Jim Stephan and Tom Kwiatkowski and David Reid and Arunkumar Byravan and Assaf Hurwitz Michaely and Nicolas Heess and Luowei Zhou and Sonam Goenka and Viral Carpenter and Anselm Levskaya and Bo Wang and Reed Roberts and Rémi Leblond and Sharat Chikkerur and Stav Ginzburg and Max Chang and Robert Riachi and Chuqiao and Xu and Zalán Borsos and Michael Pliskin and Julia Pawar and Morgane Lustman and Hannah Kirkwood and Ankit Anand and Aditi Chaudhary and Norbert Kalb and Kieran Milan and Sean Augenstein and Anna Goldie and Laurel Prince and Karthik Raman and Yanhua Sun and Vivian Xia and Aaron Cohen and Zhouyuan Huo and Josh Camp and Seher Ellis and Lukas Zilka and David Vilar Torres and Lisa Patel and Sho Arora and Betty Chan and Jonas Adler and Kareem Ayoub and Jacky Liang and Fayaz Jamil and Jiepu Jiang and Simon Baumgartner and Haitian Sun and Yael Karov and Yaroslav Akulov and Hui Zheng and Irene Cai and Claudio Fantacci and James Rubin and Alex Rav Acha and Mengchao Wang and Nina D'Souza and Rohit Sathyanarayana and Shengyang Dai and Simon Rowe and Andrey Simanovsky and Omer Goldman and Yuheng Kuang and Xiaoyue Pan and Andrew Rosenberg and Tania Rojas-Esponda and Praneet Dutta and Amy Zeng and Irina Jurenka and Greg Farquhar and Yamini Bansal and Shariq Iqbal and Becca Roelofs and Ga-Young Joung and Parker Beak and Changwan Ryu and Ryan Poplin and Yan Wu and Jean-Baptiste Alayrac and Senaka Buthpitiya and Olaf Ronneberger and Caleb Habtegebriel and Wei Li and Paul Cavallaro and Aurora Wei and Guy Bensky and Timo Denk and Harish Ganapathy and Jeff Stanway and Pratik Joshi and Francesco Bertolini and Jessica Lo and Olivia Ma and Zachary Charles and Geta Sampemane and Himanshu Sahni and Xu Chen and Harry Askham and David Gaddy and Peter Young and Jiewen Tan and Matan Eyal and Arthur Bražinskas and Li Zhong and Zhichun Wu and Mark Epstein and Kai Bailey and Andrew Hard and Kamyu Lee and Sasha Goldshtein and Alex Ruiz and Mohammed Badawi and Matthias Lochbrunner and JK Kearns and Ashley Brown and Fabio Pardo and Theophane Weber and Haichuan Yang and Pan-Pan Jiang and Berkin Akin and Zhao Fu and Marcus Wainwright and Chi Zou and Meenu Gaba and Pierre-Antoine Manzagol and Wendy Kan and Yang Song and Karina Zainullina and Rui Lin and Jeongwoo Ko and Salil Deshmukh and Apoorv Jindal and James Svensson and Divya Tyam and Heri Zhao and Christine Kaeser-Chen and Scott Baird and Pooya Moradi and Jamie Hall and Qiuchen Guo and Vincent Tsang and Bowen Liang and Fernando Pereira and Suhas Ganesh and Ivan Korotkov and Jakub Adamek and Sridhar Thiagarajan and Vinh Tran and Charles Chen and Chris Tar and Sanil Jain and Ishita Dasgupta and Taylan Bilal and David Reitter and Kai Zhao and Giulia Vezzani and Yasmin Gehman and Pulkit Mehta and Lauren Beltrone and Xerxes Dotiwalla and Sergio Guadarrama and Zaheer Abbas and Stefani Karp and Petko Georgiev and Chun-Sung Ferng and Marc Brockschmidt and Liqian Peng and Christoph Hirnschall and Vikas Verma and Yingying Bi and Ying Xiao and Avigail Dabush and Kelvin Xu and Phil Wallis and Randall Parker and Qifei Wang and Yang Xu and Ilkin Safarli and Dinesh Tewari and Yin Zhang and Seungyeon Kim and Andrea Gesmundo and Mackenzie Thomas and Sergey Levi and Ahmed Chowdhury and Kanishka Rao and Peter Garst and Sam Conway-Rahman and Helen Ran and Kay McKinney and Zhisheng Xiao and Wenhao Yu and Rohan Agrawal and Axel Stjerngren and Catalin Ionescu and Jingjing Chen and Vivek Sharma and Justin Chiu and Fei Liu and Ken Franko and Clayton Sanford and Xingyu Cai and Paul Michel and Sanjay Ganapathy and Jane Labanowski and Zachary Garrett and Ben Vargas and Sean Sun and Bryan Gale and Thomas Buschmann and Guillaume Desjardins and Nimesh Ghelani and Palak Jain and Mudit Verma and Chulayuth Asawaroengchai and Julian Eisenschlos and Jitendra Harlalka and Hideto Kazawa and Don Metzler and Joshua Howland and Ying Jian and Jake Ades and Viral Shah and Tynan Gangwani and Seungji Lee and Roman Ring and Steven M. Hernandez and Dean Reich and Amer Sinha and Ashutosh Sathe and Joe Kovac and Ashleah Gill and Ajay Kannan and Andrea D'olimpio and Martin Sevenich and Jay Whang and Been Kim and Khe Chai Sim and Jilin Chen and Jiageng Zhang and Shuba Lall and Yossi Matias and Bill Jia and Abe Friesen and Sara Nasso and Ashish Thapliyal and Bryan Perozzi and Ting Yu and Anna Shekhawat and Safeen Huda and Peter Grabowski and Eric Wang and Ashwin Sreevatsa and Hilal Dib and Mehadi Hassen and Parker Schuh and Vedrana Milutinovic and Chris Welty and Michael Quinn and Ali Shah and Bangju Wang and Gabe Barth-Maron and Justin Frye and Natalie Axelsson and Tao Zhu and Yukun Ma and Irene Giannoumis and Hanie Sedghi and Chang Ye and Yi Luan and Kevin Aydin and Bilva Chandra and Vivek Sampathkumar and Ronny Huang and Victor Lavrenko and Ahmed Eleryan and Zhi Hong and Steven Hansen and Sara Mc Carthy and Bidisha Samanta and Domagoj Ćevid and Xin Wang and Fangtao Li and Michael Voznesensky and Matt Hoffman and Andreas Terzis and Vikash Sehwag and Gil Fidel and Luheng He and Mu Cai and Yanzhang He and Alex Feng and Martin Nikoltchev and Samrat Phatale and Jason Chase and Rory Lawton and Ming Zhang and Tom Ouyang and Manuel Tragut and Mehdi Hafezi Manshadi and Arjun Narayanan and Jiaming Shen and Xu Gao and Tolga Bolukbasi and Nick Roy and Xin Li and Daniel Golovin and Liviu Panait and Zhen Qin and Guangxing Han and Thomas Anthony and Sneha Kudugunta and Viorica Patraucean and Aniket Ray and Xinyun Chen and Xiaochen Yang and Tanuj Bhatia and Pranav Talluri and Alex Morris and Andrija Ražnatović and Bethanie Brownfield and James An and Sheng Peng and Patrick Kane and Ce Zheng and Nico Duduta and Joshua Kessinger and James Noraky and Siqi Liu and Keran Rong and Petar Veličković and Keith Rush and Alex Goldin and Fanny Wei and Shiva Mohan Reddy Garlapati and Caroline Pantofaru and Okwan Kwon and Jianmo Ni and Eric Noland and Julia Di Trapani and Françoise Beaufays and Abhijit Guha Roy and Yinlam Chow and Aybuke Turker and Geoffrey Cideron and Lantao Mei and Jon Clark and Qingyun Dou and Matko Bošnjak and Ralph Leith and Yuqing Du and Amir Yazdanbakhsh and Milad Nasr and Chester Kwak and Suraj Satishkumar Sheth and Alex Kaskasoli and Ankesh Anand and Balaji Lakshminarayanan and Sammy Jerome and David Bieber and Chun-Te Chu and Alexandre Senges and Tianxiao Shen and Mukund Sridhar and Ndaba Ndebele and Benjamin Beyret and Shakir Mohamed and Mia Chen and Markus Freitag and Jiaxian Guo and Luyang Liu and Paul Roit and Heng Chen and Shen Yan and Tom Stone and JD Co-Reyes and Jeremy Cole and Salvatore Scellato and Shekoofeh Azizi and Hadi Hashemi and Alicia Jin and Anand Iyer and Marcella Valentine and András György and Arun Ahuja and Daniel Hernandez Diaz and Chen-Yu Lee and Nathan Clement and Weize Kong and Drew Garmon and Ishaan Watts and Kush Bhatia and Khyatti Gupta and Matt Miecnikowski and Hugo Vallet and Ankur Taly and Edward Loper and Saket Joshi and James Atwood and Jo Chick and Mark Collier and Fotis Iliopoulos and Ryan Trostle and Beliz Gunel and Ramiro Leal-Cavazos and Arnar Mar Hrafnkelsson and Michael Guzman and Xiaoen Ju and Andy Forbes and Jesse Emond and Kushal Chauhan and Ben Caine and Li Xiao and Wenjun Zeng and Alexandre Moufarek and Daniel Murphy and Maya Meng and Nitish Gupta and Felix Riedel and Anil Das and Elijah Lawal and Shashi Narayan and Tiberiu Sosea and James Swirhun and Linda Friso and Behnam Neyshabur and Jing Lu and Sertan Girgin and Michael Wunder and Edouard Yvinec and Aroonalok Pyne and Victor Carbune and Shruti Rijhwani and Yang Guo and Tulsee Doshi and Anton Briukhov and Max Bain and Ayal Hitron and Xuanhui Wang and Ashish Gupta and Ke Chen and Cosmo Du and Weiyang Zhang and Dhruv Shah and Arjun Akula and Max Dylla and Ashyana Kachra and Weicheng Kuo and Tingting Zou and Lily Wang and Luyao Xu and Jifan Zhu and Justin Snyder and Sachit Menon and Orhan Firat and Igor Mordatch and Yuan Yuan and Natalia Ponomareva and Rory Blevins and Lawrence Moore and Weijun Wang and Phil Chen and Martin Scholz and Artur Dwornik and Jason Lin and Sicheng Li and Diego Antognini and Te I and Xiaodan Song and Matt Miller and Uday Kalra and Adam Raveret and Oscar Akerlund and Felix Wu and Andrew Nystrom and Namrata Godbole and Tianqi Liu and Hannah DeBalsi and Jewel Zhao and Buhuang Liu and Avi Caciularu and Lauren Lax and Urvashi Khandelwal and Victoria Langston and Eric Bailey and Silvio Lattanzi and Yufei Wang and Neel Kovelamudi and Sneha Mondal and Guru Guruganesh and Nan Hua and Ofir Roval and Paweł Wesołowski and Rishikesh Ingale and Jonathan Halcrow and Tim Sohn and Christof Angermueller and Bahram Raad and Eli Stickgold and Eva Lu and Alec Kosik and Jing Xie and Timothy Lillicrap and Austin Huang and Lydia Lihui Zhang and Dominik Paulus and Clement Farabet and Alex Wertheim and Bing Wang and Rishabh Joshi and Chu-ling Ko and Yonghui Wu and Shubham Agrawal and Lily Lin and XiangHai Sheng and Peter Sung and Tyler Breland-King and Christina Butterfield and Swapnil Gawde and Sumeet Singh and Qiao Zhang and Raj Apte and Shilpa Shetty and Adrian Hutter and Tao Li and Elizabeth Salesky and Federico Lebron and Jonni Kanerva and Michela Paganini and Arthur Nguyen and Rohith Vallu and Jan-Thorsten Peter and Sarmishta Velury and David Kao and Jay Hoover and Anna Bortsova and Colton Bishop and Shoshana Jakobovits and Alessandro Agostini and Alekh Agarwal and Chang Liu and Charles Kwong and Sasan Tavakkol and Ioana Bica and Alex Greve and Anirudh GP and Jake Marcus and Le Hou and Tom Duerig and Rivka Moroshko and Dave Lacey and Andy Davis and Julien Amelot and Guohui Wang and Frank Kim and Theofilos Strinopoulos and Hui Wan and Charline Le Lan and Shankar Krishnan and Haotian Tang and Peter Humphreys and Junwen Bai and Idan Heimlich Shtacher and Diego Machado and Chenxi Pang and Ken Burke and Dangyi Liu and Renga Aravamudhan and Yue Song and Ed Hirst and Abhimanyu Singh and Brendan Jou and Liang Bai and Francesco Piccinno and Chuyuan Kelly Fu and Robin Alazard and Barak Meiri and Daniel Winter and Charlie Chen and Mingda Zhang and Jens Heitkaemper and John Lambert and Jinhyuk Lee and Alexander Frömmgen and Sergey Rogulenko and Pranav Nair and Paul Niemczyk and Anton Bulyenov and Bibo Xu and Hadar Shemtov and Morteza Zadimoghaddam and Serge Toropov and Mateo Wirth and Hanjun Dai and Sreenivas Gollapudi and Daniel Zheng and Alex Kurakin and Chansoo Lee and Kalesha Bullard and Nicolas Serrano and Ivana Balazevic and Yang Li and Johan Schalkwyk and Mark Murphy and Mingyang Zhang and Kevin Sequeira and Romina Datta and Nishant Agrawal and Charles Sutton and Nithya Attaluri and Mencher Chiang and Wael Farhan and Gregory Thornton and Kate Lin and Travis Choma and Hung Nguyen and Kingshuk Dasgupta and Dirk Robinson and Iulia Comşa and Michael Riley and Arjun Pillai and Basil Mustafa and Ben Golan and Amir Zandieh and Jean-Baptiste Lespiau and Billy Porter and David Ross and Sujeevan Rajayogam and Mohit Agarwal and Subhashini Venugopalan and Bobak Shahriari and Qiqi Yan and Hao Xu and Taylor Tobin and Pavel Dubov and Hongzhi Shi and Adrià Recasens and Anton Kovsharov and Sebastian Borgeaud and Lucio Dery and Shanthal Vasanth and Elena Gribovskaya and Linhai Qiu and Mahdis Mahdieh and Wojtek Skut and Elizabeth Nielsen and CJ Zheng and Adams Yu and Carrie Grimes Bostock and Shaleen Gupta and Aaron Archer and Chris Rawles and Elinor Davies and Alexey Svyatkovskiy and Tomy Tsai and Yoni Halpern and Christian Reisswig and Bartek Wydrowski and Bo Chang and Joan Puigcerver and Mor Hazan Taege and Jian Li and Eva Schnider and Xinjian Li and Dragos Dena and Yunhan Xu and Umesh Telang and Tianze Shi and Heiga Zen and Kyle Kastner and Yeongil Ko and Neesha Subramaniam and Aviral Kumar and Pete Blois and Zhuyun Dai and John Wieting and Yifeng Lu and Yoel Zeldes and Tian Xie and Anja Hauth and Alexandru Ţifrea and Yuqi Li and Sam El-Husseini and Dan Abolafia and Howard Zhou and Wen Ding and Sahra Ghalebikesabi and Carlos Guía and Andrii Maksai and Ágoston Weisz and Sercan Arik and Nick Sukhanov and Aga Świetlik and Xuhui Jia and Luo Yu and Weiyue Wang and Mark Brand and Dawn Bloxwich and Sean Kirmani and Zhe Chen and Alec Go and Pablo Sprechmann and Nithish Kannen and Alen Carin and Paramjit Sandhu and Isabel Edkins and Leslie Nooteboom and Jai Gupta and Loren Maggiore and Javad Azizi and Yael Pritch and Pengcheng Yin and Mansi Gupta and Danny Tarlow and Duncan Smith and Desi Ivanov and Mohammad Babaeizadeh and Ankita Goel and Satish Kambala and Grace Chu and Matej Kastelic and Michelle Liu and Hagen Soltau and Austin Stone and Shivani Agrawal and Min Kim and Kedar Soparkar and Srinivas Tadepalli and Oskar Bunyan and Rachel Soh and Arvind Kannan and DY Kim and Blake JianHang Chen and Afief Halumi and Sudeshna Roy and Yulong Wang and Olcan Sercinoglu and Gena Gibson and Sijal Bhatnagar and Motoki Sano and Daniel von Dincklage and Qingchun Ren and Blagoj Mitrevski and Mirek Olšák and Jennifer She and Carl Doersch and Jilei and Wang and Bingyuan Liu and Qijun Tan and Tamar Yakar and Tris Warkentin and Alex Ramirez and Carl Lebsack and Josh Dillon and Rajiv Mathews and Tom Cobley and Zelin Wu and Zhuoyuan Chen and Jon Simon and Swaroop Nath and Tara Sainath and Alexei Bendebury and Ryan Julian and Bharath Mankalale and Daria Ćurko and Paulo Zacchello and Adam R. Brown and Kiranbir Sodhia and Heidi Howard and Sergi Caelles and Abhinav Gupta and Gareth Evans and Anna Bulanova and Lesley Katzen and Roman Goldenberg and Anton Tsitsulin and Joe Stanton and Benoit Schillings and Vitaly Kovalev and Corey Fry and Rushin Shah and Kuo Lin and Shyam Upadhyay and Cheng Li and Soroush Radpour and Marcello Maggioni and Jing Xiong and Lukas Haas and Jenny Brennan and Aishwarya Kamath and Nikolay Savinov and Arsha Nagrani and Trevor Yacovone and Ryan Kappedal and Kostas Andriopoulos and Li Lao and YaGuang Li and Grigory Rozhdestvenskiy and Kazuma Hashimoto and Andrew Audibert and Sophia Austin and Daniel Rodriguez and Anian Ruoss and Garrett Honke and Deep Karkhanis and Xi Xiong and Qing Wei and James Huang and Zhaoqi Leng and Vittal Premachandran and Stan Bileschi and Georgios Evangelopoulos and Thomas Mensink and Jay Pavagadhi and Denis Teplyashin and Paul Chang and Linting Xue and Garrett Tanzer and Sally Goldman and Kaushal Patel and Shixin Li and Jeremy Wiesner and Ivy Zheng and Ian Stewart-Binks and Jie Han and Zhi Li and Liangchen Luo and Karel Lenc and Mario Lučić and Fuzhao Xue and Ryan Mullins and Alexey Guseynov and Chung-Ching Chang and Isaac Galatzer-Levy and Adam Zhang and Garrett Bingham and Grace Hu and Ale Hartman and Yue Ma and Jordan Griffith and Alex Irpan and Carey Radebaugh and Summer Yue and Lijie Fan and Victor Ungureanu and Christina Sorokin and Hannah Teufel and Peiran Li and Rohan Anil and Dimitris Paparas and Todd Wang and Chu-Cheng Lin and Hui Peng and Megan Shum and Goran Petrovic and Demetra Brady and Richard Nguyen and Klaus Macherey and Zhihao Li and Harman Singh and Madhavi Yenugula and Mariko Iinuma and Xinyi Chen and Kavya Kopparapu and Alexey Stern and Shachi Dave and Chandu Thekkath and Florence Perot and Anurag Kumar and Fangda Li and Yang Xiao and Matthew Bilotti and Mohammad Hossein Bateni and Isaac Noble and Lisa Lee and Amelio Vázquez-Reina and Julian Salazar and Xiaomeng Yang and Boyu Wang and Ela Gruzewska and Anand Rao and Sindhu Raghuram and Zheng Xu and Eyal Ben-David and Jieru Mei and Sid Dalmia and Zhaoyi Zhang and Yuchen Liu and Gagan Bansal and Helena Pankov and Steven Schwarcz and Andrea Burns and Christine Chan and Sumit Sanghai and Ricky Liang and Ethan Liang and Antoine He and Amy Stuart and Arun Narayanan and Yukun Zhu and Christian Frank and Bahar Fatemi and Amit Sabne and Oran Lang and Indro Bhattacharya and Shane Settle and Maria Wang and Brendan McMahan and Andrea Tacchetti and Livio Baldini Soares and Majid Hadian and Serkan Cabi and Timothy Chung and Nikita Putikhin and Gang Li and Jeremy Chen and Austin Tarango and Henryk Michalewski and Mehran Kazemi and Hussain Masoom and Hila Sheftel and Rakesh Shivanna and Archita Vadali and Ramona Comanescu and Doug Reid and Joss Moore and Arvind Neelakantan and Michaël Sander and Jonathan Herzig and Aviv Rosenberg and Mostafa Dehghani and JD Choi and Michael Fink and Reid Hayes and Eric Ge and Shitao Weng and Chia-Hua Ho and John Karro and Kalpesh Krishna and Lam Nguyen Thiet and Amy Skerry-Ryan and Daniel Eppens and Marco Andreetto and Navin Sarma and Silvano Bonacina and Burcu Karagol Ayan and Megha Nawhal and Zhihao Shan and Mike Dusenberry and Shantanu Thakoor and Sagar Gubbi and Duc Dung Nguyen and Reut Tsarfaty and Samuel Albanie and Jovana Mitrović and Meet Gandhi and Bo-Juen Chen and Alessandro Epasto and Georgi Stephanov and Ye Jin and Samuel Gehman and Aida Amini and Jack Weber and Feryal Behbahani and Shawn Xu and Miltos Allamanis and Xi Chen and Myle Ott and Claire Sha and Michal Jastrzebski and Hang Qi and David Greene and Xinyi Wu and Abodunrinwa Toki and Daniel Vlasic and Jane Shapiro and Ragha Kotikalapudi and Zhe Shen and Takaaki Saeki and Sirui Xie and Albin Cassirer and Shikhar Bharadwaj and Tatsuya Kiyono and Srinadh Bhojanapalli and Elan Rosenfeld and Sam Ritter and Jieming Mao and João Gabriel Oliveira and Zoltan Egyed and Bernd Bandemer and Emilio Parisotto and Keisuke Kinoshita and Juliette Pluto and Petros Maniatis and Steve Li and Yaohui Guo and Golnaz Ghiasi and Jean Tarbouriech and Srimon Chatterjee and Julie Jin and Katrina and Xu and Jennimaria Palomaki and Séb Arnold and Madhavi Sewak and Federico Piccinini and Mohit Sharma and Ben Albrecht and Sean Purser-haskell and Ashwin Vaswani and Chongyan Chen and Matheus Wisniewski and Qin Cao and John Aslanides and Nguyet Minh Phu and Maximilian Sieb and Lauren Agubuzu and Anne Zheng and Daniel Sohn and Marco Selvi and Anders Andreassen and Krishan Subudhi and Prem Eruvbetine and Oliver Woodman and Tomas Mery and Sebastian Krause and Xiaoqi Ren and Xiao Ma and Jincheng Luo and Dawn Chen and Wei Fan and Henry Griffiths and Christian Schuler and Alice Li and Shujian Zhang and Jean-Michel Sarr and Shixin Luo and Riccardo Patana and Matthew Watson and Dani Naboulsi and Michael Collins and Sailesh Sidhwani and Emiel Hoogeboom and Sharon Silver and Emily Caveness and Xiaokai Zhao and Mikel Rodriguez and Maxine Deines and Libin Bai and Patrick Griffin and Marco Tagliasacchi and Emily Xue and Spandana Raj Babbula and Bo Pang and Nan Ding and Gloria Shen and Elijah Peake and Remi Crocker and Shubha Srinivas Raghvendra and Danny Swisher and Woohyun Han and Richa Singh and Ling Wu and Vladimir Pchelin and Tsendsuren Munkhdalai and Dana Alon and Geoff Bacon and Efren Robles and Jannis Bulian and Melvin Johnson and George Powell and Felipe Tiengo Ferreira and Yaoyiran Li and Frederik Benzing and Mihajlo Velimirović and Hubert Soyer and William Kong and Tony and Nguyên and Zhen Yang and Jeremiah Liu and Joost van Amersfoort and Daniel Gillick and Baochen Sun and Nathalie Rauschmayr and Katie Zhang and Serena Zhan and Tao Zhou and Alexey Frolov and Chengrun Yang and Denis Vnukov and Louis Rouillard and Hongji Li and Amol Mandhane and Nova Fallen and Rajesh Venkataraman and Clara Huiyi Hu and Jennifer Brennan and Jenny Lee and Jerry Chang and Martin Sundermeyer and Zhufeng Pan and Rosemary Ke and Simon Tong and Alex Fabrikant and William Bono and Jindong Gu and Ryan Foley and Yiran Mao and Manolis Delakis and Dhruva Bhaswar and Roy Frostig and Nick Li and Avital Zipori and Cath Hope and Olga Kozlova and Swaroop Mishra and Josip Djolonga and Craig Schiff and Majd Al Merey and Eleftheria Briakou and Peter Morgan and Andy Wan and Avinatan Hassidim and RJ Skerry-Ryan and Kuntal Sengupta and Mary Jasarevic and Praveen Kallakuri and Paige Kunkle and Hannah Brennan and Tom Lieber and Hassan Mansoor and Julian Walker and Bing Zhang and Annie Xie and Goran Žužić and Adaeze Chukwuka and Alex Druinsky and Donghyun Cho and Rui Yao and Ferjad Naeem and Shiraz Butt and Eunyoung Kim and Zhipeng Jia and Mandy Jordan and Adam Lelkes and Mark Kurzeja and Sophie Wang and James Zhao and Andrew Over and Abhishek Chakladar and Marcel Prasetya and Neha Jha and Sriram Ganapathy and Yale Cong and Prakash Shroff and Carl Saroufim and Sobhan Miryoosefi and Mohamed Hammad and Tajwar Nasir and Weijuan Xi and Yang Gao and Young Maeng and Ben Hora and Chin-Yi Cheng and Parisa Haghani and Yoad Lewenberg and Caden Lu and Martin Matysiak and Naina Raisinghani and Huiyu Wang and Lexi Baugher and Rahul Sukthankar and Minh Giang and John Schultz and Noah Fiedel and Minmin Chen and Cheng-Chun Lee and Tapomay Dey and Hao Zheng and Shachi Paul and Celine Smith and Andy Ly and Yicheng Wang and Rishabh Bansal and Bartek Perz and Susanna Ricco and Stasha Blank and Vaishakh Keshava and Deepak Sharma and Marvin Chow and Kunal Lad and Komal Jalan and Simon Osindero and Craig Swanson and Jacob Scott and Anastasija Ilić and Xiaowei Li and Siddhartha Reddy Jonnalagadda and Afzal Shama Soudagar and Yan Xiong and Bat-Orgil Batsaikhan and Daniel Jarrett and Naveen Kumar and Maulik Shah and Matt Lawlor and Austin Waters and Mark Graham and Rhys May and Sabela Ramos and Sandra Lefdal and Zeynep Cankara and Nacho Cano and Brendan O'Donoghue and Jed Borovik and Frederick Liu and Jordan Grimstad and Mahmoud Alnahlawi and Katerina Tsihlas and Tom Hudson and Nikolai Grigorev and Yiling Jia and Terry Huang and Tobenna Peter Igwe and Sergei Lebedev and Xiaodan Tang and Igor Krivokon and Frankie Garcia and Melissa Tan and Eric Jia and Peter Stys and Shikhar Vashishth and Yu Liang and Balaji Venkatraman and Chenjie Gu and Anastasios Kementsietsidis and Chen Zhu and Junehyuk Jung and Yunfei Bai and Mohammad Javad Hosseini and Faruk Ahmed and Aditya Gupta and Xin Yuan and Shereen Ashraf and Shitij Nigam and Gautam Vasudevan and Pranjal Awasthi and Adi Mayrav Gilady and Zelda Mariet and Ramy Eskander and Haiguang Li and Hexiang Hu and Guillermo Garrido and Philippe Schlattner and George Zhang and Rohun Saxena and Petar Dević and Kritika Muralidharan and Ashwin Murthy and Yiqian Zhou and Min Choi and Arissa Wongpanich and Zhengdong Wang and Premal Shah and Yuntao Xu and Yiling Huang and Stephen Spencer and Alice Chen and James Cohan and Junjie Wang and Jonathan Tompson and Junru Wu and Ruba Haroun and Haiqiong Li and Blanca Huergo and Fan Yang and Tongxin Yin and James Wendt and Michael Bendersky and Rahma Chaabouni and Javier Snaider and Johan Ferret and Abhishek Jindal and Tara Thompson and Andrew Xue and Will Bishop and Shubham Milind Phal and Archit Sharma and Yunhsuan Sung and Prabakar Radhakrishnan and Mo Shomrat and Reeve Ingle and Roopali Vij and Justin Gilmer and Mihai Dorin Istin and Sam Sobell and Yang Lu and Emily Nottage and Dorsa Sadigh and Jeremiah Willcock and Tingnan Zhang and Steve Xu and Sasha Brown and Katherine Lee and Gary Wang and Yun Zhu and Yi Tay and Cheolmin Kim and Audrey Gutierrez and Abhanshu Sharma and Yongqin Xian and Sungyong Seo and Claire Cui and Elena Pochernina and Cip Baetu and Krzysztof Jastrzębski and Mimi Ly and Mohamed Elhawaty and Dan Suh and Eren Sezener and Pidong Wang and Nancy Yuen and George Tucker and Jiahao Cai and Zuguang Yang and Cindy Wang and Alex Muzio and Hai Qian and Jae Yoo and Derek Lockhart and Kevin R. McKee and Mandy Guo and Malika Mehrotra and Artur Mendonça and Sanket Vaibhav Mehta and Sherry Ben and Chetan Tekur and Jiaqi Mu and Muye Zhu and Victoria Krakovna and Hongrae Lee and AJ Maschinot and Sébastien Cevey and HyunJeong Choe and Aijun Bai and Hansa Srinivasan and Derek Gasaway and Nick Young and Patrick Siegler and Dan Holtmann-Rice and Vihari Piratla and Kate Baumli and Roey Yogev and Alex Hofer and Hado van Hasselt and Svetlana Grant and Yuri Chervonyi and David Silver and Andrew Hogue and Ayushi Agarwal and Kathie Wang and Preeti Singh and Four Flynn and Josh Lipschultz and Robert David and Lizzetth Bellot and Yao-Yuan Yang and Long Le and Filippo Graziano and Kate Olszewska and Kevin Hui and Akanksha Maurya and Nikos Parotsidis and Weijie Chen and Tayo Oguntebi and Joe Kelley and Anirudh Baddepudi and Johannes Mauerer and Gregory Shaw and Alex Siegman and Lin Yang and Shravya Shetty and Subhrajit Roy and Yunting Song and Wojciech Stokowiec and Ryan Burnell and Omkar Savant and Robert Busa-Fekete and Jin Miao and Samrat Ghosh and Liam MacDermed and Phillip Lippe and Mikhail Dektiarev and Zach Behrman and Fabian Mentzer and Kelvin Nguyen and Meng Wei and Siddharth Verma and Chris Knutsen and Sudeep Dasari and Zhipeng Yan and Petr Mitrichev and Xingyu Wang and Virat Shejwalkar and Jacob Austin and Srinivas Sunkara and Navneet Potti and Yan Virin and Christian Wright and Gaël Liu and Oriana Riva and Etienne Pot and Greg Kochanski and Quoc Le and Gargi Balasubramaniam and Arka Dhar and Yuguo Liao and Adam Bloniarz and Divyansh Shukla and Elizabeth Cole and Jong Lee and Sheng Zhang and Sushant Kafle and Siddharth Vashishtha and Parsa Mahmoudieh and Grace Chen and Raphael Hoffmann and Pranesh Srinivasan and Agustin Dal Lago and Yoav Ben Shalom and Zi Wang and Michael Elabd and Anuj Sharma and Junhyuk Oh and Suraj Kothawade and Maigo Le and Marianne Monteiro and Shentao Yang and Kaiz Alarakyia and Robert Geirhos and Diana Mincu and Håvard Garnes and Hayato Kobayashi and Soroosh Mariooryad and Kacper Krasowiak and Zhixin and Lai and Shibl Mourad and Mingqiu Wang and Fan Bu and Ophir Aharoni and Guanjie Chen and Abhimanyu Goyal and Vadim Zubov and Ankur Bapna and Elahe Dabir and Nisarg Kothari and Kay Lamerigts and Nicola De Cao and Jeremy Shar and Christopher Yew and Nitish Kulkarni and Dre Mahaarachchi and Mandar Joshi and Zhenhai Zhu and Jared Lichtarge and Yichao Zhou and Hannah Muckenhirn and Vittorio Selo and Oriol Vinyals and Peter Chen and Anthony Brohan and Vaibhav Mehta and Sarah Cogan and Ruth Wang and Ty Geri and Wei-Jen Ko and Wei Chen and Fabio Viola and Keshav Shivam and Lisa Wang and Madeleine Clare Elish and Raluca Ada Popa and Sébastien Pereira and Jianqiao Liu and Raphael Koster and Donnie Kim and Gufeng Zhang and Sayna Ebrahimi and Partha Talukdar and Yanyan Zheng and Petra Poklukar and Ales Mikhalap and Dale Johnson and Anitha Vijayakumar and Mark Omernick and Matt Dibb and Ayush Dubey and Qiong Hu and Apurv Suman and Vaibhav Aggarwal and Ilya Kornakov and Fei Xia and Wing Lowe and Alexey Kolganov and Ted Xiao and Vitaly Nikolaev and Steven Hemingray and Bonnie Li and Joana Iljazi and Mikołaj Rybiński and Ballie Sandhu and Peggy Lu and Thang Luong and Rodolphe Jenatton and Vineetha Govindaraj and Hui and Li and Gabriel Dulac-Arnold and Wonpyo Park and Henry Wang and Abhinit Modi and Jean Pouget-Abadie and Kristina Greller and Rahul Gupta and Robert Berry and Prajit Ramachandran and Jinyu Xie and Liam McCafferty and Jianling Wang and Kilol Gupta and Hyeontaek Lim and Blaž Bratanič and Andy Brock and Ilia Akolzin and Jim Sproch and Dan Karliner and Duhyeon Kim and Adrian Goedeckemeyer and Noam Shazeer and Cordelia Schmid and Daniele Calandriello and Parul Bhatia and Krzysztof Choromanski and Ceslee Montgomery and Dheeru Dua and Ana Ramalho and Helen King and Yue Gao and Lynn Nguyen and David Lindner and Divya Pitta and Oleaser Johnson and Khalid Salama and Diego Ardila and Michael Han and Erin Farnese and Seth Odoom and Ziyue Wang and Xiangzhuo Ding and Norman Rink and Ray Smith and Harshal Tushar Lehri and Eden Cohen and Neera Vats and Tong He and Parthasarathy Gopavarapu and Adam Paszke and Miteyan Patel and Wouter Van Gansbeke and Lucia Loher and Luis Castro and Maria Voitovich and Tamara von Glehn and Nelson George and Simon Niklaus and Zach Eaton-Rosen and Nemanja Rakićević and Erik Jue and Sagi Perel and Carrie Zhang and Yuval Bahat and Angéline Pouget and Zhi Xing and Fantine Huot and Ashish Shenoy and Taylor Bos and Vincent Coriou and Bryan Richter and Natasha Noy and Yaqing Wang and Santiago Ontanon and Siyang Qin and Gleb Makarchuk and Demis Hassabis and Zhuowan Li and Mandar Sharma and Kumaran Venkatesan and Iurii Kemaev and Roxanne Daniel and Shiyu Huang and Saloni Shah and Octavio Ponce and Warren and Chen and Manaal Faruqui and Jialin Wu and Slavica Andačić and Szabolcs Payrits and Daniel McDuff and Tom Hume and Yuan Cao and MH Tessler and Qingze Wang and Yinan Wang and Ivor Rendulic and Eirikur Agustsson and Matthew Johnson and Tanya Lando and Andrew Howard and Sri Gayatri Sundara Padmanabhan and Mayank Daswani and Andrea Banino and Michael Kilgore and Jonathan Heek and Ziwei Ji and Alvaro Caceres and Conglong Li and Nora Kassner and Alexey Vlaskin and Zeyu Liu and Alex Grills and Yanhan Hou and Roykrong Sukkerd and Gowoon Cheon and Nishita Shetty and Larisa Markeeva and Piotr Stanczyk and Tejas Iyer and Yuan Gong and Shawn Gao and Keerthana Gopalakrishnan and Tim Blyth and Malcolm Reynolds and Avishkar Bhoopchand and Misha Bilenko and Dero Gharibian and Vicky Zayats and Aleksandra Faust and Abhinav Singh and Min Ma and Hongyang Jiao and Sudheendra Vijayanarasimhan and Lora Aroyo and Vikas Yadav and Sarah Chakera and Ashwin Kakarla and Vilobh Meshram and Karol Gregor and Gabriela Botea and Evan Senter and Dawei Jia and Geza Kovacs and Neha Sharma and Sebastien Baur and Kai Kang and Yifan He and Lin Zhuo and Marija Kostelac and Itay Laish and Songyou Peng and Louis O'Bryan and Daniel Kasenberg and Girish Ramchandra Rao and Edouard Leurent and Biao Zhang and Sage Stevens and Ana Salazar and Ye Zhang and Ivan Lobov and Jake Walker and Allen Porter and Morgan Redshaw and Han Ke and Abhishek Rao and Alex Lee and Hoi Lam and Michael Moffitt and Jaeyoun Kim and Siyuan Qiao and Terry Koo and Robert Dadashi and Xinying Song and Mukund Sundararajan and Peng Xu and Chizu Kawamoto and Yan Zhong and Clara Barbu and Apoorv Reddy and Mauro Verzetti and Leon Li and George Papamakarios and Hanna Klimczak-Plucińska and Mary Cassin and Koray Kavukcuoglu and Rigel Swavely and Alain Vaucher and Jeffrey Zhao and Ross Hemsley and Michael Tschannen and Heming Ge and Gaurav Menghani and Yang Yu and Natalie Ha and Wei He and Xiao Wu and Maggie Song and Rachel Sterneck and Stefan Zinke and Dan A. Calian and Annie Marsden and Alejandro Cruzado Ruiz and Matteo Hessel and Almog Gueta and Benjamin Lee and Brian Farris and Manish Gupta and Yunjie Li and Mohammad Saleh and Vedant Misra and Kefan Xiao and Piermaria Mendolicchio and Gavin Buttimore and Varvara Krayvanova and Nigamaa Nayakanti and Matthew Wiethoff and Yash Pande and Azalia Mirhoseini and Ni Lao and Jasmine Liu and Yiqing Hua and Angie Chen and Yury Malkov and Dmitry Kalashnikov and Shubham Gupta and Kartik Audhkhasi and Yuexiang Zhai and Sudhindra Kopalle and Prateek Jain and Eran Ofek and Clemens Meyer and Khuslen Baatarsukh and Hana Strejček and Jun Qian and James Freedman and Ricardo Figueira and Michal Sokolik and Olivier Bachem and Raymond Lin and Dia Kharrat and Chris Hidey and Pingmei Xu and Dennis Duan and Yin Li and Muge Ersoy and Richard Everett and Kevin Cen and Rebeca Santamaria-Fernandez and Amir Taubenfeld and Ian Mackinnon and Linda Deng and Polina Zablotskaia and Shashank Viswanadha and Shivanker Goel and Damion Yates and Yunxiao Deng and Peter Choy and Mingqing Chen and Abhishek Sinha and Alex Mossin and Yiming Wang and Arthur Szlam and Susan Hao and Paul Kishan Rubenstein and Metin Toksoz-Exley and Miranda Aperghis and Yin Zhong and Junwhan Ahn and Michael Isard and Olivier Lacombe and Florian Luisier and Chrysovalantis Anastasiou and Yogesh Kalley and Utsav Prabhu and Emma Dunleavy and Shaan Bijwadia and Justin Mao-Jones and Kelly Chen and Rama Pasumarthi and Emily Wood and Adil Dostmohamed and Nate Hurley and Jiri Simsa and Alicia Parrish and Mantas Pajarskas and Matt Harvey and Ondrej Skopek and Yony Kochinski and Javier Rey and Verena Rieser and Denny Zhou and Sun Jae Lee and Trilok Acharya and Guowang Li and Joe Jiang and Xiaofan Zhang and Bryant Gipson and Ethan Mahintorabi and Marco Gelmi and Nima Khajehnouri and Angel Yeh and Kayi Lee and Loic Matthey and Leslie Baker and Trang Pham and Han Fu and Alex Pak and Prakhar Gupta and Cristina Vasconcelos and Adam Sadovsky and Brian Walker and Sissie Hsiao and Patrik Zochbauer and Andreea Marzoca and Noam Velan and Junhao Zeng and Gilles Baechler and Danny Driess and Divya Jain and Yanping Huang and Lizzie Tao and John Maggs and Nir Levine and Jon Schneider and Erika Gemzer and Samuel Petit and Shan Han and Zach Fisher and Dustin Zelle and Courtney Biles and Eugene Ie and Asya Fadeeva and Casper Liu and Juliana Vicente Franco and Adrian Collister and Hao Zhang and Renshen Wang and Ruizhe Zhao and Leandro Kieliger and Kurt Shuster and Rui Zhu and Boqing Gong and Lawrence Chan and Ruoxi Sun and Sujoy Basu and Roland Zimmermann and Jamie Hayes and Abhishek Bapna and Jasper Snoek and Weel Yang and Puranjay Datta and Jad Al Abdallah and Kevin Kilgour and Lu Li and SQ Mah and Yennie Jun and Morgane Rivière and Abhijit Karmarkar and Tammo Spalink and Tao Huang and Lucas Gonzalez and Duc-Hieu Tran and Averi Nowak and John Palowitch and Martin Chadwick and Ellie Talius and Harsh Mehta and Thibault Sellam and Philipp Fränken and Massimo Nicosia and Kyle He and Aditya Kini and David Amos and Sugato Basu and Harrison Jobe and Eleni Shaw and Qiantong Xu and Colin Evans and Daisuke Ikeda and Chaochao Yan and Larry Jin and Lun Wang and Sachin Yadav and Ilia Labzovsky and Ramesh Sampath and Ada Ma and Candice Schumann and Aditya Siddhant and Rohin Shah and John Youssef and Rishabh Agarwal and Natalie Dabney and Alessio Tonioni and Moran Ambar and Jing Li and Isabelle Guyon and Benny Li and David Soergel and Boya Fang and Georgi Karadzhov and Cristian Udrescu and Trieu Trinh and Vikas Raunak and Seb Noury and Dee Guo and Sonal Gupta and Mara Finkelstein and Denis Petek and Lihao Liang and Greg Billock and Pei Sun and David Wood and Yiwen Song and Xiaobin Yu and Tatiana Matejovicova and Regev Cohen and Kalyan Andra and David D'Ambrosio and Zhiwei Deng and Vincent Nallatamby and Ebrahim Songhori and Rumen Dangovski and Andrew Lampinen and Pankil Botadra and Adam Hillier and Jiawei Cao and Nagabhushan Baddi and Adhi Kuncoro and Toshihiro Yoshino and Ankit Bhagatwala and Marcáurelio Ranzato and Rylan Schaeffer and Tianlin Liu and Shuai Ye and Obaid Sarvana and John Nham and Chenkai Kuang and Isabel Gao and Jinoo Baek and Shubham Mittal and Ayzaan Wahid and Anita Gergely and Bin Ni and Josh Feldman and Carrie Muir and Pascal Lamblin and Wolfgang Macherey and Ethan Dyer and Logan Kilpatrick and Víctor Campos and Mukul Bhutani and Stanislav Fort and Yanif Ahmad and Aliaksei Severyn and Kleopatra Chatziprimou and Oleksandr Ferludin and Mason Dimarco and Aditya Kusupati and Joe Heyward and Dan Bahir and Kevin Villela and Katie Millican and Dror Marcus and Sanaz Bahargam and Caglar Unlu and Nicholas Roth and Zichuan Wei and Siddharth Gopal and Deepanway Ghoshal and Edward Lee and Sharon Lin and Jennie Lees and Dayeong Lee and Anahita Hosseini and Connie Fan and Seth Neel and Marcus Wu and Yasemin Altun and Honglong Cai and Enrique Piqueras and Josh Woodward and Alessandro Bissacco and Salem Haykal and Mahyar Bordbar and Prasha Sundaram and Sarah Hodkinson and Daniel Toyama and George Polovets and Austin Myers and Anu Sinha and Tomer Levinboim and Kashyap Krishnakumar and Rachita Chhaparia and Tatiana Sholokhova and Nitesh Bharadwaj Gundavarapu and Ganesh Jawahar and Haroon Qureshi and Jieru Hu and Nikola Momchev and Matthew Rahtz and Renjie Wu and Aishwarya P S and Kedar Dhamdhere and Meiqi Guo and Umang Gupta and Ali Eslami and Mariano Schain and Michiel Blokzijl and David Welling and Dave Orr and Levent Bolelli and Nicolas Perez-Nieves and Mikhail Sirotenko and Aman Prasad and Arjun Kar and Borja De Balle Pigem and Tayfun Terzi and Gellért Weisz and Dipankar Ghosh and Aditi Mavalankar and Dhruv Madeka and Kaspar Daugaard and Hartwig Adam and Viraj Shah and Dana Berman and Maggie Tran and Steven Baker and Ewa Andrejczuk and Grishma Chole and Ganna Raboshchuk and Mahdi Mirzazadeh and Thais Kagohara and Shimu Wu and Christian Schallhart and Bernett Orlando and Chen Wang and Alban Rrustemi and Hao Xiong and Hao Liu and Arpi Vezer and Nolan Ramsden and Shuo-yiin Chang and Sidharth Mudgal and Yan Li and Nino Vieillard and Yedid Hoshen and Farooq Ahmad and Ambrose Slone and Amy Hua and Natan Potikha and Mirko Rossini and Jon Stritar and Sushant Prakash and Zifeng Wang and Xuanyi Dong and Alireza Nazari and Efrat Nehoran and Kaan Tekelioglu and Yinxiao Li and Kartikeya Badola and Tom Funkhouser and Yuanzhen Li and Varun Yerram and Ramya Ganeshan and Daniel Formoso and Karol Langner and Tian Shi and Huijian Li and Yumeya Yamamori and Amayika Panda and Alaa Saade and Angelo Scorza Scarpati and Chris Breaux and CJ Carey and Zongwei Zhou and Cho-Jui Hsieh and Sophie Bridgers and Alena Butryna and Nishesh Gupta and Vaibhav Tulsyan and Sanghyun Woo and Evgenii Eltyshev and Will Grathwohl and Chanel Parks and Seth Benjamin and Rina Panigrahy and Shenil Dodhia and Daniel De Freitas and Chris Sauer and Will Song and Ferran Alet and Jackson Tolins and Cosmin Paduraru and Xingyi Zhou and Brian Albert and Zizhao Zhang and Lei Shu and Mudit Bansal and Sarah Nguyen and Amir Globerson and Owen Xiao and James Manyika and Tom Hennigan and Rong Rong and Josip Matak and Anton Bakalov and Ankur Sharma and Danila Sinopalnikov and Andrew Pierson and Stephen Roller and Geoff Brown and Mingcen Gao and Toshiyuki Fukuzawa and Amin Ghafouri and Kenny Vassigh and Iain Barr and Zhicheng Wang and Anna Korsun and Rajesh Jayaram and Lijie Ren and Tim Zaman and Samira Khan and Yana Lunts and Dan Deutsch and Dave Uthus and Nitzan Katz and Masha Samsikova and Amr Khalifa and Nikhil Sethi and Jiao Sun and Luming Tang and Uri Alon and Xianghong Luo and Dian Yu and Abhishek Nayyar and Bryce Petrini and Will Truong and Vincent Hellendoorn and Nikolai Chinaev and Chris Alberti and Wei Wang and Jingcao Hu and Vahab Mirrokni and Ananth Balashankar and Avia Aharon and Aahil Mehta and Ahmet Iscen and Joseph Kready and Lucas Manning and Anhad Mohananey and Yuankai Chen and Anshuman Tripathi and Allen Wu and Igor Petrovski and Dawsen Hwang and Martin Baeuml and Shreyas Chandrakaladharan and Yuan Liu and Rey Coaguila and Maxwell Chen and Sally Ma and Pouya Tafti and Susheel Tatineni and Terry Spitz and Jiayu Ye and Paul Vicol and Mihaela Rosca and Adrià Puigdomènech and Zohar Yahav and Sanjay Ghemawat and Hanzhao Lin and Phoebe Kirk and Zaid Nabulsi and Sergey Brin and Bernd Bohnet and Ken Caluwaerts and Aditya Srikanth Veerubhotla and Dan Zheng and Zihang Dai and Petre Petrov and Yichong Xu and Ramin Mehran and Zhuo Xu and Luisa Zintgraf and Jiho Choi and Spurthi Amba Hombaiah and Romal Thoppilan and Sashank Reddi and Lukasz Lew and Li Li and Kellie Webster and KP Sawhney and Lampros Lamprou and Siamak Shakeri and Mayank Lunayach and Jianmin Chen and Sumit Bagri and Alex Salcianu and Ying Chen and Yani Donchev and Charlotte Magister and Signe Nørly and Vitor Rodrigues and Tomas Izo and Hila Noga and Joe Zou and Thomas Köppe and Wenxuan Zhou and Kenton Lee and Xiangzhu Long and Danielle Eisenbud and Anthony Chen and Connor Schenck and Chi Ming To and Peilin Zhong and Emanuel Taropa and Minh Truong and Omer Levy and Danilo Martins and Zhiyuan Zhang and Christopher Semturs and Kelvin Zhang and Alex Yakubovich and Pol Moreno and Lara McConnaughey and Di Lu and Sam Redmond and Lotte Weerts and Yonatan Bitton and Tiziana Refice and Nicolas Lacasse and Arthur Conmy and Corentin Tallec and Julian Odell and Hannah Forbes-Pollard and Arkadiusz Socala and Jonathan Hoech and Pushmeet Kohli and Alanna Walton and Rui Wang and Mikita Sazanovich and Kexin Zhu and Andrei Kapishnikov and Rich Galt and Matthew Denton and Ben Murdoch and Caitlin Sikora and Kareem Mohamed and Wei Wei and Uri First and Tim McConnell and Luis C. Cobo and James Qin and Thi Avrahami and Daniel Balle and Yu Watanabe and Annie Louis and Adam Kraft and Setareh Ariafar and Yiming Gu and Eugénie Rives and Charles Yoon and Andrei Rusu and James Cobon-Kerr and Chris Hahn and Jiaming Luo and Yuvein and Zhu and Niharika Ahuja and Rodrigo Benenson and Raphaël Lopez Kaufman and Honglin Yu and Lloyd Hightower and Junlin Zhang and Darren Ni and Lisa Anne Hendricks and Gabby Wang and Gal Yona and Lalit Jain and Pablo Barrio and Surya Bhupatiraju and Siva Velusamy and Allan Dafoe and Sebastian Riedel and Tara Thomas and Zhe Yuan and Mathias Bellaiche and Sheena Panthaplackel and Klemen Kloboves and Sarthak Jauhari and Canfer Akbulut and Todor Davchev and Evgeny Gladchenko and David Madras and Aleksandr Chuklin and Tyrone Hill and Quan Yuan and Mukundan Madhavan and Luke Leonhard and Dylan Scandinaro and Qihang Chen and Ning Niu and Arthur Douillard and Bogdan Damoc and Yasumasa Onoe and Fabian Pedregosa and Fred Bertsch and Chas Leichner and Joseph Pagadora and Jonathan Malmaud and Sameera Ponda and Andy Twigg and Oleksii Duzhyi and Jingwei Shen and Miaosen Wang and Roopal Garg and Jing Chen and Utku Evci and Jonathan Lee and Leon Liu and Koji Kojima and Masa Yamaguchi and Arunkumar Rajendran and AJ Piergiovanni and Vinodh Kumar Rajendran and Marco Fornoni and Gabriel Ibagon and Harry Ragan and Sadh MNM Khan and John Blitzer and Andrew Bunner and Guan Sun and Takahiro Kosakai and Scott Lundberg and Ndidi Elue and Kelvin Guu and SK Park and Jane Park and Arunachalam Narayanaswamy and Chengda Wu and Jayaram Mudigonda and Trevor Cohn and Hairong Mu and Ravi Kumar and Laura Graesser and Yichi Zhang and Richard Killam and Vincent Zhuang and Mai Giménez and Wael Al Jishi and Ruy Ley-Wild and Alex Zhai and Kazuki Osawa and Diego Cedillo and Jialu Liu and Mayank Upadhyay and Marcin Sieniek and Roshan Sharma and Tom Paine and Anelia Angelova and Sravanti Addepalli and Carolina Parada and Kingshuk Majumder and Avery Lamp and Sanjiv Kumar and Xiang Deng and Artiom Myaskovsky and Tea Sabolić and Jeffrey Dudek and Sarah York and Félix de Chaumont Quitry and Jiazhong Nie and Dee Cattle and Alok Gunjan and Bilal Piot and Waleed Khawaja and Seojin Bang and Simon Wang and Siavash Khodadadeh and Raghavender R and Praynaa Rawlani and Richard Powell and Kevin Lee and Johannes Griesser and GS Oh and Cesar Magalhaes and Yujia Li and Simon Tokumine and Hadas Natalie Vogel and Dennis Hsu and Arturo BC and Disha Jindal and Matan Cohen and Zi Yang and Junwei Yuan and Dario de Cesare and Tony Bruguier and Jun Xu and Monica Roy and Alon Jacovi and Dan Belov and Rahul Arya and Phoenix Meadowlark and Shlomi Cohen-Ganor and Wenting Ye and Patrick Morris-Suzuki and Praseem Banzal and Gan Song and Pranavaraj Ponnuramu and Fred Zhang and George Scrivener and Salah Zaiem and Alif Raditya Rochman and Kehang Han and Badih Ghazi and Kate Lee and Shahar Drath and Daniel Suo and Antonious Girgis and Pradeep Shenoy and Duy Nguyen and Douglas Eck and Somit Gupta and Le Yan and Joao Carreira and Anmol Gulati and Ruoxin Sang and Daniil Mirylenka and Emma Cooney and Edward Chou and Mingyang Ling and Cindy Fan and Ben Coleman and Guilherme Tubone and Ravin Kumar and Jason Baldridge and Felix Hernandez-Campos and Angeliki Lazaridou and James Besley and Itay Yona and Neslihan Bulut and Quentin Wellens and AJ Pierigiovanni and Jasmine George and Richard Green and Pu Han and Connie Tao and Geoff Clark and Chong You and Abbas Abdolmaleki and Justin Fu and Tongzhou Chen and Ashwin Chaugule and Angad Chandorkar and Altaf Rahman and Will Thompson and Penporn Koanantakool and Mike Bernico and Jie Ren and Andrey Vlasov and Sergei Vassilvitskii and Maciej Kula and Yizhong Liang and Dahun Kim and Yangsibo Huang and Chengxi Ye and Dmitry Lepikhin and Wesley Helmholz},
      year={2025},
      eprint={2507.06261},
      archivePrefix={arXiv},
      primaryClass={cs.CL},
      url={https://arxiv.org/abs/2507.06261}, 
}

@misc{deepseekai2025deepseekv32,
      title={DeepSeek-V3.2: Pushing the Frontier of Open Large Language Models}, 
      author={DeepSeek-AI},
      year={2025},
}

@misc{gemini3pro2025modelcard,
  title = {Gemini 3 Pro Model Card},
  author = {{Gemini Team, Google}},
  year = {2025},
  url = {https://storage.googleapis.com/deepmind-media/Model-Cards/Gemini-3-Pro-Model-Card.pdf},
  institution = {Google DeepMind},
}

@inproceedings{dong2024survey,
    title = "A Survey on In-context Learning",
    author = "Dong, Qingxiu  and
      Li, Lei  and
      Dai, Damai  and
      Zheng, Ce  and
      Ma, Jingyuan  and
      Li, Rui  and
      Xia, Heming  and
      Xu, Jingjing  and
      Wu, Zhiyong  and
      Chang, Baobao  and
      Sun, Xu  and
      Li, Lei  and
      Sui, Zhifang",
    editor = "Al-Onaizan, Yaser  and
      Bansal, Mohit  and
      Chen, Yun-Nung",
    booktitle = "Proceedings of the 2024 Conference on Empirical Methods in Natural Language Processing",
    month = nov,
    year = "2024",
    address = "Miami, Florida, USA",
    publisher = "Association for Computational Linguistics",
    url = "https://aclanthology.org/2024.emnlp-main.64/",
    doi = "10.18653/v1/2024.emnlp-main.64",
    pages = "1107--1128"
}

@article{luo2025large,
  title={Large language model agent: A survey on methodology, applications and challenges},
  author={Luo, Junyu and Zhang, Weizhi and Yuan, Ye and Zhao, Yusheng and Yang, Junwei and Gu, Yiyang and Wu, Bohan and Chen, Binqi and Qiao, Ziyue and Long, Qingqing and others},
  journal={arXiv preprint arXiv:2503.21460},
  year={2025}
}

@article{ruijters2015fault,
  title={Fault tree analysis: A survey of the state-of-the-art in modeling, analysis and tools},
  author={Ruijters, Enno and Stoelinga, Mari{\"e}lle},
  journal={Computer science review},
  volume={15},
  pages={29--62},
  year={2015},
  publisher={Elsevier}
}

@article{liu2023visual,
  title={Visual instruction tuning},
  author={Liu, Haotian and Li, Chunyuan and Wu, Qingyang and Lee, Yong Jae},
  journal={Advances in neural information processing systems},
  volume={36},
  pages={34892--34916},
  year={2023}
}

@misc{zhao2023survey,
      title={A Survey of Large Language Models}, 
      author={Wayne Xin Zhao and Kun Zhou and Junyi Li and Tianyi Tang and Xiaolei Wang and Yupeng Hou and Yingqian Min and Beichen Zhang and Junjie Zhang and Zican Dong and Yifan Du and Chen Yang and Yushuo Chen and Zhipeng Chen and Jinhao Jiang and Ruiyang Ren and Yifan Li and Xinyu Tang and Zikang Liu and Peiyu Liu and Jian-Yun Nie and Ji-Rong Wen},
      year={2026},
      eprint={2303.18223},
      archivePrefix={arXiv},
      primaryClass={cs.CL},
      url={https://arxiv.org/abs/2303.18223}, 
}

@article{lee2009fault,
  title={Fault tree analysis, methods, and applications a review},
  author={Lee, Wen-Shing and Grosh, Doris L and Tillman, Frank A and Lie, Chang H},
  journal={IEEE transactions on reliability},
  volume={34},
  number={3},
  pages={194--203},
  year={2009},
  publisher={IEEE}
}

@techreport{vesely1981fault,
  title={Fault tree handbook},
  author={Vesely, William E and Goldberg, Francine F and Roberts, Norman H and Haasl, David F},
  year={1981}
}

@inproceedings{liu2023deplot,
    title = "{D}e{P}lot: One-shot visual language reasoning by plot-to-table translation",
    author = "Liu, Fangyu  and
      Eisenschlos, Julian  and
      Piccinno, Francesco  and
      Krichene, Syrine  and
      Pang, Chenxi  and
      Lee, Kenton  and
      Joshi, Mandar  and
      Chen, Wenhu  and
      Collier, Nigel  and
      Altun, Yasemin",
    editor = "Rogers, Anna  and
      Boyd-Graber, Jordan  and
      Okazaki, Naoaki",
    booktitle = "Findings of the Association for Computational Linguistics: ACL 2023",
    month = jul,
    year = "2023",
    address = "Toronto, Canada",
    publisher = "Association for Computational Linguistics",
    url = "https://aclanthology.org/2023.findings-acl.660/",
    doi = "10.18653/v1/2023.findings-acl.660",
    pages = "10381--10399"
}

@misc{feng2025vision,
      title={Vision Remember: Recovering Visual Information in Efficient LVLM with Vision Feature Resampling}, 
      author={Ze Feng and Jiang-jiang Liu and Sen Yang and Lingyu Xiao and Zhibin Quan and Zhenhua Feng and Wankou Yang and Jingdong Wang},
      year={2025},
      eprint={2506.03928},
      archivePrefix={arXiv},
      primaryClass={cs.CV},
      url={https://arxiv.org/abs/2506.03928}, 
}

@inproceedings{zhang2025pfdial,
    title = "{PFD}ial: A Structured Dialogue Instruction Fine-tuning Method Based on {UML} Flowcharts",
    author = "Zhang, Ming  and
      Wang, Yuhui  and
      Shen, Yujiong  and
      Yang, Tingyi  and
      Jiang, Changhao  and
      Wu, Yilong  and
      Dou, Shihan  and
      Chen, Qinhao  and
      Xi, Zhiheng  and
      Zhang, Zhihao  and
      Dong, Yi  and
      Wang, Zhen  and
      Fei, Zhihui  and
      Wan, Mingyang  and
      Liang, Tao  and
      Ma, Guojun  and
      Zhang, Qi  and
      Gui, Tao  and
      Huang, Xuanjing",
    editor = "Che, Wanxiang  and
      Nabende, Joyce  and
      Shutova, Ekaterina  and
      Pilehvar, Mohammad Taher",
    booktitle = "Findings of the Association for Computational Linguistics: ACL 2025",
    month = jul,
    year = "2025",
    address = "Vienna, Austria",
    publisher = "Association for Computational Linguistics",
    url = "https://aclanthology.org/2025.findings-acl.134/",
    doi = "10.18653/v1/2025.findings-acl.134",
    pages = "2626--2649",
    ISBN = "979-8-89176-256-5"
}

@inproceedings{ericson1999fault,
  title={Fault tree analysis},
  author={Ericson, Clifton A and Ll, Clifton},
  booktitle={System Safety Conference, Orlando, Florida},
  volume={1},
  pages={1--9},
  year={1999}
}

@article{eckberg1963ws,
  title={WS-133B Fault tree analysis program plan},
  author={Eckberg, CR},
  journal={Seattle, Washington: The Boeing Company},
  year={1963}
}

@techreport{goldberg1994system,
  title={System engineering toolbox for design-oriented engineers},
  author={Goldberg, Benjamin E and Everhart, K and Stevens, R and Babbitt III, N and Clemens, P and Stout, L},
  year={1994}
}

@book{american1985guidelines,
  title={Guidelines for Hazard Evaluation Procedures},
  author={American Institute of Chemical Engineers. Center for Chemical Process Safety},
  year={1985},
  publisher={American Institute of Chemical Engineers}
}

@article{vcepin2002dynamic,
  title={A dynamic fault tree},
  author={{\v{C}}epin, Marko and Mavko, Borut},
  journal={Reliability Engineering \& System Safety},
  volume={75},
  number={1},
  pages={83--91},
  year={2002},
  publisher={Elsevier}
}

@article{iec2006fault,
  title={Fault Tree Analysis (FTA)},
  author={IEC 61025 Technical Committee and others},
  journal={IEC 61025},
  year={2006},
  publisher={IEC}
}

@misc{wang2023survey,
      title={A Survey of the Evolution of Language Model-Based Dialogue Systems: Data, Task and Models}, 
      author={Hongru Wang and Lingzhi Wang and Yiming Du and Liang Chen and Jingyan Zhou and Yufei Wang and Kam-Fai Wong},
      year={2025},
      eprint={2311.16789},
      archivePrefix={arXiv},
      primaryClass={cs.CL},
      url={https://arxiv.org/abs/2311.16789}, 
}

@misc{budzianowski2018multiwoz,
      title={MultiWOZ -- A Large-Scale Multi-Domain Wizard-of-Oz Dataset for Task-Oriented Dialogue Modelling}, 
      author={Paweł Budzianowski and Tsung-Hsien Wen and Bo-Hsiang Tseng and Iñigo Casanueva and Stefan Ultes and Osman Ramadan and Milica Gašić},
      year={2020},
      eprint={1810.00278},
      archivePrefix={arXiv},
      primaryClass={cs.CL},
      url={https://arxiv.org/abs/1810.00278}, 
}

@article{zhu2020crosswoz,
  title={Crosswoz: A large-scale chinese cross-domain task-oriented dialogue dataset},
  author={Zhu, Qi and Huang, Kaili and Zhang, Zheng and Zhu, Xiaoyan and Huang, Minlie},
  journal={Transactions of the Association for Computational Linguistics},
  volume={8},
  pages={281--295},
  year={2020},
  publisher={MIT Press One Rogers Street, Cambridge, MA 02142-1209, USA journals-info~…}
}

@article{wang2020kddres,
  title={KddRES: A multi-level Knowledge-driven dialogue dataset for restaurant towards customized dialogue system},
  author={Wang, Hongru and Li, Min and Zhou, Zimo and Fung, Gabriel Pui Cheong and Wong, Kam-Fai},
  journal={arXiv preprint arXiv:2011.08772},
  year={2020}
}

@inproceedings{zhang2024transfertod,
  title={Transfertod: A generalizable chinese multi-domain task-oriented dialogue system with transfer capabilities},
  author={Zhang, Ming and Huang, Caishuang and Wu, Yilong and Liu, Shichun and Zheng, Huiyuan and Dong, Yurui and Shen, Yujiong and Dou, Shihan and Zhao, Jun and Ye, Junjie and others},
  booktitle={Proceedings of the 2024 Conference on Empirical Methods in Natural Language Processing},
  pages={12750--12771},
  year={2024}
}

@inproceedings{kwan2024mt,
  title={MT-Eval: A Multi-Turn Capabilities Evaluation Benchmark for Large Language Models},
  author={Kwan, Wai-Chung and Zeng, Xingshan and Jiang, Yuxin and Wang, Yufei and Li, Liangyou and Shang, Lifeng and Jiang, Xin and Liu, Qun and Wong, Kam-Fai},
  booktitle={Proceedings of the 2024 Conference on Empirical Methods in Natural Language Processing},
  pages={20153--20177},
  year={2024}
}

@inproceedings{bai2024mt,
  title={MT-Bench-101: A Fine-Grained Benchmark for Evaluating Large Language Models in Multi-Turn Dialogues},
  author={Bai, Ge and Liu, Jie and Bu, Xingyuan and He, Yancheng and Liu, Jiaheng and Zhou, Zhanhui and Lin, Zhuoran and Su, Wenbo and Ge, Tiezheng and Zheng, Bo and others},
  booktitle={Proceedings of the 62nd Annual Meeting of the Association for Computational Linguistics (Volume 1: Long Papers)},
  pages={7421--7454},
  year={2024}
}

@article{yehudai2025survey,
  title={Survey on evaluation of llm-based agents},
  author={Yehudai, Asaf and Eden, Lilach and Li, Alan and Uziel, Guy and Zhao, Yilun and Bar-Haim, Roy and Cohan, Arman and Shmueli-Scheuer, Michal},
  journal={arXiv preprint arXiv:2503.16416},
  year={2025}
}

@article{yao2024tau,
  title={$tau $-bench: A Benchmark for Tool-Agent-User Interaction in Real-World Domains},
  author={Yao, Shunyu and Shinn, Noah and Razavi, Pedram and Narasimhan, Karthik},
  journal={arXiv preprint arXiv:2406.12045},
  year={2024}
}

@inproceedings{huang2025crmarena,
  title={Crmarena: Understanding the capacity of llm agents to perform professional crm tasks in realistic environments},
  author={Huang, Kung-Hsiang and Prabhakar, Akshara and Dhawan, Sidharth and Mao, Yixin and Wang, Huan and Savarese, Silvio and Xiong, Caiming and Laban, Philippe and Wu, Chien-Sheng},
  booktitle={Proceedings of the 2025 Conference of the Nations of the Americas Chapter of the Association for Computational Linguistics: Human Language Technologies (Volume 1: Long Papers)},
  pages={3830--3850},
  year={2025}
}

@article{wang2023mint,
  title={Mint: Evaluating llms in multi-turn interaction with tools and language feedback},
  author={Wang, Xingyao and Wang, Zihan and Liu, Jiateng and Chen, Yangyi and Yuan, Lifan and Peng, Hao and Ji, Heng},
  journal={arXiv preprint arXiv:2309.10691},
  year={2023}
}

@inproceedings{patilberkeley,
  title={The Berkeley Function Calling Leaderboard (BFCL): From Tool Use to Agentic Evaluation of Large Language Models},
  author={Patil, Shishir G and Mao, Huanzhi and Yan, Fanjia and Ji, Charlie Cheng-Jie and Suresh, Vishnu and Stoica, Ion and Gonzalez, Joseph E},
  year=2025,
  booktitle={Forty-second International Conference on Machine Learning}
}

@inproceedings{tang2025basic,
  title={BASIC: Boosting Visual Alignment with Intrinsic Refined Embeddings in Multimodal Large Language Models},
  author={Tang, Jianting and Wang, Yubo and Cao, Haoyu and Xu, Linli},
  booktitle={Proceedings of the IEEE/CVF International Conference on Computer Vision},
  pages={20582--20592},
  year={2025}
}

@article{hurst2024gpt,
  title={Gpt-4o system card},
  author={Hurst, Aaron and Lerer, Adam and Goucher, Adam P and Perelman, Adam and Ramesh, Aditya and Clark, Aidan and Ostrow, AJ and Welihinda, Akila and Hayes, Alan and Radford, Alec and others},
  journal={arXiv preprint arXiv:2410.21276},
  year={2024}
}

@techreport{anthropic2025claude45,
  title       = {Claude 4.5 Sonnet System Card},
  author      = {Anthropic},
  institution = {Anthropic PBC},
  year        = {2025},
  url         = {https://assets.anthropic.com/m/12f214efcc2f457a/original/Claude-Sonnet-4-5-System-Card.pdf},
  note        = {Technical Report}
}

@inproceedings{yao2022react,
  title={React: Synergizing reasoning and acting in language models},
  author={Yao, Shunyu and Zhao, Jeffrey and Yu, Dian and Du, Nan and Shafran, Izhak and Narasimhan, Karthik R and Cao, Yuan},
  booktitle={The eleventh international conference on learning representations},
  year={2022}
}

@article{yang2025qwen3,
  title={Qwen3 technical report},
  author={Yang, An and Li, Anfeng and Yang, Baosong and Zhang, Beichen and Hui, Binyuan and Zheng, Bo and Yu, Bowen and Gao, Chang and Huang, Chengen and Lv, Chenxu and others},
  journal={arXiv preprint arXiv:2505.09388},
  year={2025}
}

@article{comanici2025gemini,
  title={Gemini 2.5: Pushing the frontier with advanced reasoning, multimodality, long context, and next generation agentic capabilities},
  author={Comanici, Gheorghe and Bieber, Eric and Schaekermann, Mike and Pasupat, Ice and Sachdeva, Noveen and Dhillon, Inderjit and Blistein, Marcel and Ram, Ori and Zhang, Dan and Rosen, Evan and others},
  journal={arXiv preprint arXiv:2507.06261},
  year={2025}
}

@inproceedings{zheng2024llamafactory,
  title={LlamaFactory: Unified Efficient Fine-Tuning of 100+ Language Models},
  author={Yaowei Zheng and Richong Zhang and Junhao Zhang and Yanhan Ye and Zheyan Luo and Zhangchi Feng and Yongqiang Ma},
  booktitle={Proceedings of the 62nd Annual Meeting of the Association for Computational Linguistics (Volume 3: System Demonstrations)},
  address={Bangkok, Thailand},
  publisher={Association for Computational Linguistics},
  year={2024},
  url={http://arxiv.org/abs/2403.13372}
}

@article{guo2025deepseek,
  title={Deepseek-r1: Incentivizing reasoning capability in llms via reinforcement learning},
  author={Guo, Daya and Yang, Dejian and Zhang, Haowei and Song, Junxiao and Zhang, Ruoyu and Xu, Runxin and Zhu, Qihao and Ma, Shirong and Wang, Peiyi and Bi, Xiao and others},
  journal={arXiv preprint arXiv:2501.12948},
  year={2025}
}

@article{schulman2017proximal,
  title={Proximal policy optimization algorithms},
  author={Schulman, John and Wolski, Filip and Dhariwal, Prafulla and Radford, Alec and Klimov, Oleg},
  journal={arXiv preprint arXiv:1707.06347},
  year={2017}
}

@article{sheng2024hybridflow,
  title   = {HybridFlow: A Flexible and Efficient RLHF Framework},
  author  = {Guangming Sheng and Chi Zhang and Zilingfeng Ye and Xibin Wu and Wang Zhang and Ru Zhang and Yanghua Peng and Haibin Lin and Chuan Wu},
  year    = {2024},
  journal = {arXiv preprint arXiv: 2409.19256}
}

@article{shao2024deepseekmath,
  title={Deepseekmath: Pushing the limits of mathematical reasoning in open language models},
  author={Shao, Zhihong and Wang, Peiyi and Zhu, Qihao and Xu, Runxin and Song, Junxiao and Bi, Xiao and Zhang, Haowei and Zhang, Mingchuan and Li, YK and Wu, Yang and others},
  journal={arXiv preprint arXiv:2402.03300},
  year={2024}
}

\newpage
\appendix

\clearpage
\section{JFTA Details}
\label{apd:JFTA}
 
\subsection{Structural Constraints}

To ensure the validity of the representation for automated reasoning, JFTA imposes graph-theoretic constraints. The overall structure must remain acyclic to prevent infinite reasoning loops, and solution nodes must exclusively appear at the leaf level. These constraints align JFTA with the fundamental assumptions of classical FTA while facilitating efficient parsing and consistency checking.

\begin{table*}[ht]
\centering
\begin{tabular}{p{2cm} p{1.5cm} p{11.0cm}}
\toprule
\textbf{Field Name} & \textbf{Type} & \textbf{Description} \\
\midrule
ID 
& String 
& A globally unique identifier within the fault tree. \\

\midrule
NodeName 
& String 
& Natural language text describing a fault phenomenon, cause, or solution. \\

\midrule
NextType 
& Enum
& Specifies the logical type and behavior of the node, including \texttt{XOR}, \texttt{OR}, \texttt{AND}, \texttt{Fault}, \texttt{Solution}, and \texttt{LINK}. This enumeration is extensible to support additional types.\\

\midrule
NextTree 
& Object 
& When \texttt{NextType} is \texttt{XOR}, \texttt{OR}, \texttt{AND}, or \texttt{Fault}, this field is an object containing child nodes. \\

& String 
& When \texttt{NextType} is \texttt{Solution}, this field contains a textual description of the solution; when it is \texttt{LINK}, this field stores the referenced node ID. \\
\bottomrule
\end{tabular}
\caption{Fields of a JFTA Basic Node}
\label{tab:JFTA_node_fields}
\end{table*}

\subsection{Implementation detail}
In the implementation, each node is uniquely identified by an ID that serves as the key in the JSON representation, while the remaining attributes are organized as the corresponding value in a dictionary structure. Logical gate behavior is thus embedded within semantic nodes through their associated type definitions. Table~\ref{tab:JFTA_node_fields} summarizes the fields of a basic JFTA node.

\subsection{Fault Tree Examples}
\label{apd:JFTA-eg}
\begin{promptbox}[Fault Tree Examples]
{
 "1": {
  "NodeName": "Light does not turn on",
  "NextType": "XOR",
  "NextTree": {
   "2": {
    "NodeName": "Switch Closed",
    "NextType": "OR",
    "NextTree": {
     "4": {
      "NodeName": "Power Supply Issue",
      "NextType": "OR",
      "NextTree": {
       "8": {
        "NodeName": "Power not connected",
        "NextType": "Solution",
        "NextTree": "Connect power"
       },
       "9": {
        "NodeName": "Power plug loose",
        "NextType": "Solution",
        "NextTree": "Secure power plug"
       },
       "10": {
        "NodeName": "Insufficient voltage",
        "NextType": "Solution",
        "NextTree": "Install transformer"
       }
      }
     },
     "5": {
      "NodeName": "Circuit Issue",
      "NextType": "Fault",
      "NextTree": {
       "13": {
        "NodeName": "Circuit Issue (Test intermediate faults)",
        "NextType": "OR",
        "NextTree": {
         "11": {
          "NodeName": "Wire breakage",
          "NextType": "Solution",
          "NextTree": "Replace wire"
         },
         "12": {
          "NodeName": "Poor contact",
          "NextType": "Solution",
          "NextTree": "Repair contact point"
         }
        }
       }
      }
     },
     "6": {
      "NodeName": "Bulb Issue",
      "NextType": "Solution",
      "NextTree": "Replace bulb"
     },
     "7": {
      "NodeName": "Switch Broken",
      "NextType": "Solution",
      "NextTree": "Replace switch"
     },
     "14": {
      "NodeName": "Power Supply Issue (Test DAG)",
      "NextType": "LINK",
      "NextTree": "4"
     },
     "15": {
      "NodeName": "Voltage Issue (Test DAG)",
      "NextType": "LINK",
      "NextTree": "10"
     }
    }
   },
   "3": {
    "NodeName": "Switch Open",
    "NextType": "Solution",
    "NextTree": "Close switch"
   }
  }
 }
}
\end{promptbox}

\section{Dataset Details}

\subsection{Domains and scenarios}
\label{apd:domains}
The table \ref{tab:failures_part1} and table \ref{tab:failures_part2} present the domains and failure scenarios included in the data.

\begin{table*}[h!]
\centering
\begin{tabularx}{\textwidth}{>{\raggedright\arraybackslash}p{4cm} >{\raggedright\arraybackslash}X}
\toprule
\textbf{Domain} & \textbf{Failure Scenarios} \\
\midrule
Power \& Energy & Large-scale power grid blackout; Nuclear power plant safety system failure; Petrochemical plant explosion/leakage; Wind power/Photovoltaic (PV) system fault; Energy storage station fire/explosion; Smart grid data tampering. \\
\midrule
Aerospace & Aircraft engine failure; Flight control system paralysis; Spacecraft launch failure; Satellite in-orbit service failure; UAV (Drone) crash/loss of contact. \\
\midrule
Automotive \& Transportation & Automotive braking system failure; Railway signaling system fault; Subway system suspension; Ship propulsion/navigation fault; Electric Vehicle (EV) thermal management failure; High-level autonomous driving misjudgment. \\
\midrule
Medical \& Life Sciences & Ventilator/Pacemaker failure; Hospital power/oxygen supply interruption; Medical robot malfunction; In-vitro diagnostics (IVD) equipment misdiagnosis; Biobank temperature control failure. \\
\midrule
Information \& Communication & Data center downtime; Communication network interruption; Satellite communication failure; Large-scale cloud service outage; Content Delivery Network (CDN) scheduling anomaly. \\
\midrule
Construction \& Public Safety & Fire protection system failure; Elevator suspension accident; High-rise building water supply interruption; Smart building security system failure; Large stadium evacuation failure; Supercomputer center liquid cooling leak. \\
\midrule
Mining \& Industrial Safety & Gas explosion accident; Mine flooding/roof collapse; Chemical production unit failure; Industrial robot injury accident; Polymerization reactor runaway; Hazardous chemical storage chain explosion; Digital factory industrial control network paralysis. \\
\midrule
Defense \& Military & Weapon system malfunction; Military vehicle/vessel failure; Battlefield communication interruption; Reconnaissance satellite data transmission failure. \\
\midrule
Microelectronics \& Semiconductors & Chip tape-out failure; Wafer manufacturing low yield; Advanced packaging reliability failure; Integrated circuit electrostatic discharge (ESD) breakdown. \\
\midrule
Computer Systems \& Software & Operating system kernel panic; Distributed database data inconsistency; Online payment system double deduction; AI model performance degradation in production; Recommender system filter bubble reinforcement; Smart home system failure; Cloud computing resource scheduling imbalance. \\
\midrule
Emerging Tech \& Frontiers & Quantum computer fidelity drop; Gene editing off-target effects; Brain-Computer Interface (BCI) command misinterpretation; Metaverse virtual space motion sickness; Synthetic biology engineered strain escape; Digital twin vs. physical entity distortion. \\
\bottomrule
\end{tabularx}
\caption{Potential Failures in Various Domains (Part 1)}
\label{tab:failures_part1}
\end{table*}

\begin{table*}[htbp]
\centering
\begin{tabularx}{\textwidth}{>{\raggedright\arraybackslash}p{4cm} >{\raggedright\arraybackslash}X}
\toprule
\textbf{Domain} & \textbf{Failure Scenarios} \\
\midrule

Finance \& Infrastructure & Stock exchange trading system interruption; Automated Market Maker (AMM) protocol manipulation; Digital currency wallet private key loss; AI financial risk control model failure. \\
\midrule
Environment \& Ecology & Chemical plant toxic substance leak into river; Nuclear waste geological repository leakage; Carbon Capture and Storage (CCS) failure; Deep-sea aquaculture cage mass fish mortality; Chemical wastewater treatment system failure. \\
\midrule
Public Health \& Biosecurity & Pathogen laboratory leak; Public health surveillance system reporting failure; Uncontrolled community outbreak of infectious disease; Mass vaccination failure. \\
\midrule
Agriculture \& Food Engineering & Smart irrigation system failure; Precision fertilization system disorder; Greenhouse environment control loss; Livestock barn environment deterioration; Cold chain logistics break; Transgenic crop gene drift. \\
\midrule
Ocean Engineering \& Polar & Deep-sea drilling platform blowout; Wave energy device structural fatigue fracture; Polar station energy supply chain break; Autonomous Underwater Vehicle (AUV) mission failure; Submarine optical cable communication interruption; Icebreaker propulsion system disability. \\
\midrule
Sports Events & Smart referee system misjudgment; Stadium energy system failure; Event broadcast signal interruption; Ticketing system crash; Athlete monitoring equipment failure. \\
\midrule
Tourism Attractions & Scenic area passenger flow monitoring failure; Amusement ride suspension accident; Smart guide platform downtime; Artifact storage constant temp/humidity system anomaly; Emergency evacuation system failure. \\
\midrule
Education & Online teaching platform downtime; Smart exam system cheating/failure; Campus "One Card" system anomaly; Teaching building power outage; Virtual laboratory simulation error. \\
\midrule
Urban Infrastructure & City-wide traffic light paralysis; Urban heating system failure; Smart streetlight system anomaly; Underground drainage pump station failure; City emergency broadcast malfunction; Smart parking system crash. \\
\midrule
Media \& Entertainment & Large-scale event safety accident; Live streaming platform massive downtime; Sports event broadcast interruption; News gathering/editing system data loss; Short video recommendation anomaly; Digital Rights Management (DRM) crash; False news reporting. \\
\midrule
Judiciary & Electronic court system downtime; Electronic evidence chain failure; Prison security system failure; Notary office electronic signature system downtime; Judicial appraisal laboratory equipment fault. \\
\midrule
Rail Transit & Signaling system fault; Train traction system failure; Train braking system failure; Platform screen door system anomaly; Train Control System (CBTC/ATO) crash. \\
\bottomrule
\end{tabularx}
\caption{Potential Failures in Various Domains (Part 2)}
\label{tab:failures_part2}
\end{table*}


\subsection{Path Sampling Algorithm}
Algorithm \ref{alg:sampling} presents the algorithm for sampling paths by expanding siblings.

\begin{algorithm}[h!]
  \caption{Path Sampling Algorithm}
  \label{alg:sampling}
  \begin{algorithmic}
    \STATE {\bfseries Input:} Fault Tree $G$, Root $r$, Basic Solutions $S_{init}$
    \STATE {\bfseries Output:} Selected Path Nodes $V_{selected}$
    
    \STATE
    
    \STATE {\bfseries Function} \textsc{PathSampling}($G, r, S_{init}$)
    \STATE Initialize $V_{selected} \leftarrow \emptyset$
    
    \FORALL{$s \in S_{init}$}
      \STATE \textsc{TracebackToRoot}($s, r$)
    \ENDFOR
    
    \REPEAT
      \STATE $S_{new} \leftarrow$ \textsc{Expand}($V_{selected}$)
      \IF{$S_{new} \neq \emptyset$}
        \FORALL{$s \in S_{new}$}
          \STATE \textsc{TracebackToRoot}($s, r$)
        \ENDFOR
      \ENDIF
    \UNTIL{$S_{new}$ is empty}
    
    \STATE \textbf{return} \textsc{DFS}($V_{selected}$)
    
    \STATE
    
    \STATE {\bfseries Function} \textsc{TracebackToRoot}($u, r$)
    \IF{$u \in V_{selected}$}
      \STATE \textbf{return}
    \ENDIF
    \STATE $V_{selected} \leftarrow V_{selected} \cup \{u\}$
    \IF{$u \neq r$ \AND $Parents(u) \neq \emptyset$}
      \STATE $p \leftarrow$ Randomly select one from $Parents(u)$
      \STATE \textsc{TracebackToRoot}($p, r$)
    \ENDIF
    
    \STATE
    
    \STATE {\bfseries Function} \textsc{ExpandAndSiblings}($V_{curr}$)
    \STATE $S_{added} \leftarrow \emptyset$
    \FORALL{$u \in V_{curr}$}
      \IF{\textsc{NeedExpend}($u$)}
        \FORALL{child $c \in Children(u)$}
          \IF{$c \notin V_{curr}$}
            \STATE $S_{sub} \leftarrow$ \textsc{GetoneSolution}($c$)
            \STATE $S_{added} \leftarrow S_{added} \cup S_{sub} $
          \ENDIF
        \ENDFOR
      \ENDIF
    \ENDFOR
    \STATE \textbf{return} $S_{added}$
    
  \end{algorithmic}
\end{algorithm}

\section{User Simulator Details}

\subsection{User Simulation Dataset Statistics}
\label{apd:user-data-statistics}

Table~\ref{tab:user_data_appendix} presents the detailed statistics of the user simulation dataset used for training and evaluating the user simulator. 
The dataset contains three types of samples: solution confirmation, fault confirmation, and hacking samples for invalid assistant requests. 
We also report the average input and output lengths calculated by the Qwen3-8B tokenizer.

\begin{table}[htbp]
  \centering
  \begin{tabular}{lrr}
    \toprule
    \textbf{Metric} & \textbf{Training Set} & \textbf{Testing Set} \\
    \midrule
    \textbf{Total Samples} & 37,115 & 9,714 \\
    \midrule
    \textit{Category} & & \\
    \quad Solution & 13,524 & 3,566  \\
    \quad Fault & 22,808  & 6,061 \\
    \quad Hacking & 783  & 87  \\
    \midrule
    \textit{Length} & & \\
    \quad Input  & 1,296.09 & 1,301.37 \\
    \quad Output & 51.18 & 51.68 \\
    \bottomrule
  \end{tabular}
  \caption{Statistics of the training and testing datasets for the user simulator.}
  \label{tab:user_data_appendix}
\end{table}

\subsection{Human--LLM Judge Consistency}
\label{apd:judge-consistency}

To assess the reliability of the LLM-as-a-Judge evaluation used for user response quality, we conduct an additional consistency check between two human annotators and the GPT-based judge on 50 examples. 
For each user simulator, Annotator A, Annotator B, and GPT independently score the generated responses along the same two dimensions: fuzziness compliance and naturalness. 
The total score is computed as the sum of the two dimensions. 
As shown in Table~\ref{tab:judge_consistency}, although the absolute scores from human annotators differ from the GPT-4o scores, and there are minor differences in the relative ordering between GRPO and SFT, both human annotators consistently identify PPO as the best-performing method. They also agree that SFT and GRPO outperform the base model. 
This indicates that the LLM judge provides a stable and reasonably aligned approximation of human preference when comparing different user simulators.

\begin{table*}[htbp]
\centering
\small
\setlength{\tabcolsep}{5pt}
\begin{tabular}{lcccc cccc cccc}
\toprule
\multirow{2}{*}{\textbf{Annotator}}
& \multicolumn{4}{c}{\textbf{Fuzziness}}
& \multicolumn{4}{c}{\textbf{Naturalness}}
& \multicolumn{4}{c}{\textbf{Total}} \\
\cmidrule(lr){2-5} \cmidrule(lr){6-9} \cmidrule(lr){10-13}
& \textbf{Base} & \textbf{SFT} & \textbf{GRPO} & \textbf{PPO}
& \textbf{Base} & \textbf{SFT} & \textbf{GRPO} & \textbf{PPO}
& \textbf{Base} & \textbf{SFT} & \textbf{GRPO} & \textbf{PPO} \\
\midrule

Human A
& \underline{1.56} & 1.76 & \textbf{1.78} & \textbf{1.78}
& \underline{1.38} & 1.82 & 1.80 & \textbf{1.88}
& \underline{2.94} & 3.58 & 3.58 & \textbf{3.66} \\

Human B
& \underline{0.62} & 1.92 & 1.88 & \textbf{2.08}
& \underline{0.92} & 1.22 & 1.24 & \textbf{1.32}
& \underline{1.54} & 3.14 & 3.12 & \textbf{3.40} \\

GPT
& \underline{0.63} & 2.24 & 2.21 & \textbf{2.29}
& \underline{0.94} & 1.51 & 1.46 & \textbf{1.53}
& \underline{1.57} & 3.75 & 3.67 & \textbf{3.82} \\

$\Delta$
& -0.46 & +0.40 & +0.38 & +0.36
& -0.21 & -0.01 & -0.06 & -0.07
& -0.67 & +0.39 & +0.32 & +0.29 \\

\bottomrule
\end{tabular}
\caption{
Consistency check between human annotators and the GPT-based judge.
$\Delta$ denotes the difference between the GPT score and the average score of the two human annotators for the same method.
For each annotator and each evaluation dimension, the highest score among the four methods is shown in \textbf{bold}, while the lowest score is \underline{underlined}.
}
\label{tab:judge_consistency}
\end{table*}

\section{Experiment Details}
\subsection{Rollback Ablation Study}
\label{apd:rollback-ablation}

To evaluate the impact of the rollback mechanism on benchmark difficulty, we conduct an ablation study using DeepSeek-V3.2 as the assistant model. 
Specifically, we compare two evaluation settings: one without rollback, where the user consistently follows a single fault setting throughout the dialogue, and one with rollback, where the user may revise earlier observations and switch to a corrected fault setting. 
As shown in Table~\ref{tab:rollback_ablation}, adding rollback substantially reduces the task completion rate from 58.87\% to 41.40\%. 
This result indicates that rollback scenarios introduce additional challenges beyond standard fault localization, requiring the assistant to track long-range dialogue history, identify revised observations, and update its diagnostic state accordingly.

\begin{table}[htbp]
\centering
\begin{tabular}{lc}
\toprule
\textbf{Setting} & \textbf{Correct(\%)} \\
\midrule
Without rollback & 58.87 \\
With rollback & 41.40 \\
\bottomrule
\end{tabular}
\caption{
Rollback ablation study using DeepSeek-V3.2.
}
\label{tab:rollback_ablation}
\end{table}

\subsection{Representative Failure Cases}
\label{apd:failure-cases}

To better illustrate the error patterns discussed in Section~\ref{sec:analyze}, we provide representative failure cases observed during model evaluation. 
These examples show that model failures arise from incorrect traversal order, premature solution generation, inconsistency between reasoning and action, rollback failure, and repeated confirmation of the same node. For readability, all cases presented in this section are equivalent English translations of the cases in our experiments.

\subsubsection{Traversal Not Following DFS Order}

In this case, the assistant skips the expected DFS traversal order. 
According to the DFS order, node 3 should be checked before node 45. 
Only after confirming node 3 should the assistant continue to node 45. 
However, the model directly jumps from node 2 to node 45, probably because of the strong semantic association between the names of node 2 and node 45.

\begin{promptbox}[Failure Case: Non-DFS Traversal]
{
    "role": "assistant",
    "content": "{
    \"think\": \"The user reported the top-level fault 'reconnaissance satellite data cannot be transmitted back'. According to the fault tree, this fault is caused by three intermediate-level faults connected by an OR relationship. I need to start checking from the first intermediate node, 'on-board system failure'.\",
    \"ID\": \"2\",
    \"response\": \"First, check whether there is a fault in the on-board system.\"
}"
},
{
    "role": "user",
    "content": "It feels like the system responsible for overall operation is not quite right. It doesn't seem to coordinate as smoothly as usual."
},
{
    "role": "assistant",
    "content": "{
    \"think\": \"The user's description that 'the system responsible for overall operation is not quite right' points to a coordination problem inside the on-board system. Under 'on-board system failure', 'on-board processor failure' is the core component responsible for data processing and system coordination, so it needs to be checked first.\",
    \"ID\": \"45\",
    \"response\": \"It is necessary to check whether the on-board processor is working properly.\"
}"
}
\end{promptbox}

\subsubsection{Premature Solution Generation}

In this case, after identifying the problem at the penultimate layer, the model is still uncertain about the specific leaf-node fault. 
However, it directly provides a solution instead of continuing to distinguish among the child nodes. 
This causes the troubleshooting process to terminate before the actual bottom-level fault is confirmed.

\begin{promptbox}[Failure Case: Premature Solution Generation]
{
    "role": "assistant",
    "content": "{
    \"think\": \"The user confirmed 'main power supply failure' (ID: 82). This node has two child nodes: generator failure (ID: 90) and distribution board failure (ID: 91). The user's description that 'the main circuit or equipment is not operating properly' may refer either to the generator (power generation equipment) or to the distribution board (power distribution). Further distinction is needed.\",
    \"ID\": \"90\",
    \"response\": \"Please check whether the generator is operating abnormally.\"
}"
},
{
    "role": "user",
    "content": "I feel like that big piece of equipment that provides power for us is not quite right now. The sound it makes when it rotates is different from usual."
},
{
    "role": "assistant",
    "content": "{
    \"think\": \"The user confirmed 'generator failure' (ID: 90). This node has three XOR child nodes: stator winding fault, rotor fault, and excitation system fault. The user's description that 'the sound it makes when it rotates is different from usual' is usually related to mechanical rotating components, so it most likely points to 'rotor fault' (ID: 101).\",
    \"ID\": \"101\",
    \"response\": \"A rotor fault is usually resolved by repairing or replacing the rotor.\"
}"
}
\end{promptbox}

\subsubsection{Mismatch Between Reasoning and Action}

In this case, the \texttt{think} field indicates that the model should continue troubleshooting a child node. 
However, the actual response is phrased more like a conclusion or solution rather than a fault confirmation question. 
As a result, the troubleshooting logic is not correctly reflected in the model action.

\begin{promptbox}[Failure Case: Reasoning-Action Mismatch]
{
    "role": "assistant",
    "content": "{
    \"think\": \"The user corrected the judgment and confirmed the problem of 'inappropriate number of network layers'. This node has two child nodes. I need to identify the specific cause based on the user's description of 'poor coordination'. The user mentioned 'data transmission' and 'coordination', which is more inclined toward 'gradient vanishing caused by excessive depth' rather than 'insufficient feature extraction caused by insufficient depth'. Therefore, I will check 'gradient vanishing caused by excessive depth'.\",
    \"ID\": \"68\",
    \"response\": \"Poor data transmission may be caused by gradient vanishing due to an excessively deep network.\"
}"
}
\end{promptbox}

\subsubsection{Backtracking Failure After Rollback}

In this case, the user explicitly states that a previous judgment was wrong. 
The model should backtrack to the corrected node and reconsider the corresponding subtree. 
Instead, it continues along the previous path and rechecks node 47, showing that it fails to update the diagnostic state after rollback.

\begin{promptbox}[Failure Case: Backtracking Failure]
{
    "role": "assistant",
    "content": "{
    \"think\": \"The user ruled out the possibility of 'seal failure' (ID: 43). According to the OR logic, the last child node, 'blockage' (ID: 47), needs to be checked.\",
    \"ID\": \"47\",
    \"response\": \"Next, check whether there is a blockage in the pump or pipeline.\"
}"
},
{
    "role": "user",
    "content": "Sorry, I made an incorrect judgment earlier about the pump mechanical failure, and the judgments for the subsequent nodes may also be problematic. After observing again, I heard that the big device responsible for transporting liquid sounds a bit abnormal. The sound is a little muffled, and it feels like it is not rotating as smoothly as usual."
},
{
    "role": "assistant",
    "content": "{
    \"think\": \"The user corrected the judgment and described the pump's abnormal sound and unsmooth rotation again. This still points to 'pump mechanical failure' (ID: 38). Since impeller wear and seal failure were previously ruled out, the possibility of 'blockage' (ID: 47) now needs to be reconsidered, because blockage can also cause unsmooth rotation and abnormal noise.\",
    \"ID\": \"47\",
    \"response\": \"Okay, let us recheck whether there is a blockage in the pump or pipeline.\"
}"
}
\end{promptbox}

\subsubsection{Repeated Confirmation of the Same Node}

In this case, the assistant asks about the same node repeatedly. 
Although the user has already provided diagnostic feedback for node 85, the model continues to request confirmation for the same problem. 
This indicates that the model fails to effectively use dialogue history to advance the troubleshooting process.

\begin{promptbox}[Failure Case: Repeated Confirmation]
{
    "role": "assistant",
    "content": "{
    \"think\": \"The user states that the network isolation mechanism is operating normally and no abnormal connection has been found. This preliminarily rules out the possibility of 'network isolation failure' (ID: 65). Continue checking other OR child nodes under the top-level fault. Next, ask about 'engineer station misoperation' (ID: 85).\",
    \"ID\": \"85\",
    \"response\": \"Next, please check whether there is any misoperation at the engineer station.\"
}"
},
{
    "role": "user",
    "content": "I feel like on the operation interface side, some steps may be a little different from what we are used to, and it feels somewhat awkward to operate."
},
{
    "role": "assistant",
    "content": "{
    \"think\": \"The user again mentions that the operation interface and steps are different from usual habits, and that the operation is awkward. This is consistent with the previously described system abnormality, but this time it more specifically points to the possibility of 'engineer station misoperation'. Therefore, I need to further confirm whether this intermediate fault exists.\",
    \"ID\": \"85\",
    \"response\": \"Abnormal operation interface and steps may be caused by misoperation. Please confirm whether there is any misoperation at the engineer station.\"
}"
},
{
    "role": "user",
    "content": "I saw that the operator seemed to perform some actions on the interface used to monitor and control the whole system. I cannot say exactly what happened, but it feels somewhat different from the usual operation process."
},
{
    "role": "assistant",
    "content": "{
    \"think\": \"The user described that the operator performed operations at the engineer station that were different from usual, which further points to 'engineer station misoperation' (ID: 85). To move the troubleshooting process forward, I need the user to explicitly confirm or deny this fault.\",
    \"ID\": \"85\",
    \"response\": \"The operator performed actions different from the normal process, which may be misoperation. Please confirm whether there is any misoperation at the engineer station.\"
}"
},
{
    "role": "user",
    "content": "It seems that when operating the interface used for monitoring and control, some steps looked different from the usual habits. It feels a bit wrong."
}
\end{promptbox}

\subsection{Image-based Representation Ablation}
\label{apd:image-ablation}

To further examine whether fault trees can be directly understood from images by multimodal LLMs, we conduct a small-scale image-based ablation study. 
This experiment is different from the main multi-turn benchmark and is substantially easier. 
Instead of requiring the model to interact with a simulated user and recover from rollback, we provide the model with the ground-truth basic faults and ask it to identify the corresponding paths from the root node to these basic faults and provide the associated solutions.

We use a Level-2 subset for this evaluation. 
Since high-resolution fault tree images are large, with an average file size of 8.84MB, they exceed the file-size limit of our command-line submission pipeline for Gemini(7MB). 
Therefore, the controlled comparison between JFTA and image inputs is conducted using GPT-5.5. 
We additionally include DeepSeek-V3.2 under the JFTA setting as a reference. 
For each method, we report two metrics: \textit{tree pass rate}, which requires all paths in a fault tree to be correctly identified, and \textit{path pass rate}, which measures the proportion of correctly identified paths.

\begin{table*}[htbp]
\centering
\begin{tabular}{llcc}
\toprule
\textbf{Input} & \textbf{Model} & \textbf{Tree Pass Rate} & \textbf{Path Pass Rate} \\
\midrule
JFTA  & DeepSeek-V3.2 & 96.77 & 98.97 \\
JFTA  & GPT-5.5       & 100.00 & 100.00 \\
Image & GPT-5.5       & 0.00 & 0.00 \\
\bottomrule
\end{tabular}
\caption{
Small-scale comparison between JFTA and image-based fault tree inputs.
}
\label{tab:image_ablation}
\end{table*}

As shown in Table~\ref{tab:image_ablation}, GPT-5.5 achieves perfect performance when using JFTA, but fails completely when using only the fault tree image. 
This suggests that, for large and dense fault trees, current vision-language models may still struggle to reliably capture graph topology, gate types, node identities, and relative spatial relationships from images. 
By contrast, JFTA explicitly encodes these structural elements in a compact and standardized textual form, making the diagnostic path much easier for LLMs to recover.

To further inspect the image understanding capability of state-of-the-art closed-source models, we also conduct qualitative tests through the web interface with Gemini 3.1 Pro and GPT-5.5. 
Both models fail to pass the image-based test. 
Gemini explicitly indicates that it cannot reliably process the high-resolution image and then produces guessed paths with hallucinated nodes. 
GPT-5.5 attempts to use Python tools to crop, enlarge, enhance, and OCR the image, but the extracted paths remain incorrect and also contain hallucinated content. 
These observations further support our motivation: directly using fault tree images is unreliable for complex diagnostic reasoning, while even a simple node-edge textual representation can substantially improve path recovery. 
JFTA further improves over node-edge representations by providing higher information density and a standardized hierarchical structure.

\section{Hyperparameters}

\subsection{Supervised Finetuning }
\label{apd:sft}
The table \ref{tab:sft_hyperparams} presents the hyper-parameters used for Supervised Fine-Tuning.
\begin{table}[H]
    \centering
    \begin{tabular}{lc}
        \toprule
        \textbf{Hyperparameter} & \textbf{Value} \\
        \midrule
        Batch Size & 64 \\
        Learning Rate & $1.0 \times 10^{-5}$ \\
        LR Scheduler & Cosine \\
        Optimizer & AdamW  \\
        Warmup Ratio & 0.1 \\
        Epochs & 3 \\
        Precision & bf16 \\
        \bottomrule
    \end{tabular}
    \caption{Hyperparameters used for SFT.}
    \label{tab:sft_hyperparams}
\end{table}
\subsection{Reinforcement Learning}
\label{apd:ppo}
The table \ref{tab:ppo_hyperparams} presents the hyper-parameters used for PPO Training.
\begin{table}[H]
    \centering
    \begin{tabular}{lc}
        \toprule
        \textbf{Hyperparameter} & \textbf{Value} \\
        \midrule
        Total Epochs & 3 \\
        Advantage Estimator & GAE \\
        Global Train Batch Size & 256 \\
        Precision & bf16 \\
        Actor Learning Rate & $1.0 \times 10^{-6}$ \\
        PPO Mini-Batch Size & 64 \\
        KL Loss Coefficient & 0.001 \\
        Optimizer & AdamW  \\
        Critic Learning Rate & $1.0 \times 10^{-5}$ \\
        Critic Warmup & 0 \\
        Max Response Length & 1024 \\
        \bottomrule
    \end{tabular}
    \caption{Hyperparameters used for PPO Reinforcement Learning.}
    \label{tab:ppo_hyperparams}
\end{table}

\section{Prompts}
JFTA-Bench is constructed and evaluated based on a Chinese-language dataset. 
For readability, all prompts presented in this section are equivalent English translations of the prompts used in our experiments. 
The original Chinese prompts will be provided in the future code release.

\subsection{Prompt for Assistant}
\label{apd:prompt-assistant}
\begin{promptbox}[Prompt for Assistant]
You are now acting as a technical support expert in the **{field}** domain. Your task is to assist users in systematically troubleshooting faults and providing final solutions.

---

## Task Objective

The fault tree consists of three levels: top-level, intermediate-level, and bottom-level:

* Bottom-level faults are the root causes of the problem and can be directly fixed;
* Intermediate-level and top-level faults are higher-level manifestations of bottom-level faults.

The user will initially describe a top-level fault. Your task is to:

1. Start from the top-level node and conduct step-by-step troubleshooting down the fault tree;
2. Gradually identify the true root cause by confirming with the user whether a specific intermediate or bottom-level fault exists;
3. Once a bottom-level fault is confirmed, provide a clear, actionable, and engineering-feasible solution;
4. A single scenario may involve multiple bottom-level faults; only after all of them are resolved can the issue be considered fully fixed.

You will also receive the user's historical troubleshooting records and solution verification feedback to maintain consistency and continuity of the fault tree state.

Below is the fault tree structure you have access to:
{fault_tree}

---

## Output Format

Always respond using the following JSON format:

```json
{
    "think": "Based on the user's response...",
    "ID": "string",
    "response": "The troubleshooting action to perform in this round or the solution to be provided"
}
```

---

## Response Rules

### 1. Node ID Rules

The "ID" field must correspond to the **node ID actually being handled in the next turn**:

* If this round involves troubleshooting, fill in the ID of the node you are querying;
* If this round provides a solution, fill in the ID of the corresponding bottom-level fault.

It must be consistent with the operation in the next dialogue turn.

---

### 2. Assistant Response Rules

When troubleshooting or providing solutions, responses must be naturally phrased and maintain a professional technical tone, while explicitly stating the **full fault name** currently being handled. Responses should be concise (within 20 words), only describing the *object to be checked* or the *action to be taken*, without elaborating on details.

To avoid templated outputs, expressions must remain varied. Do not use the same sentence structure in two consecutive turns, and do not mechanically repeat fixed patterns such as "need to confirm whether ... fault exists."

**During the troubleshooting phase, the response must include the full fault name.**
**During the solution phase, the response must include both the full fault name and the method.**

> **Mandatory requirements**:
>
> * Do not repeat the same sentence structure within three turns.
> * Do not use fixed templates such as "XXX is usually handled by repair or replacement."
> * Do not use templated phrasing like "need to confirm whether XXX fault exists."

---

### 3. Think Field

This field is used to explain your reasoning process, including but not limited to:

* How you ruled out or confirmed a node based on the user's previous feedback;
* Why the current node should be handled in this round;
* Why it is appropriate to enter the solution phase at this point.

The thinking content does not need to repeat node IDs or solution details; it should only reflect the reasoning process.

---

## Examples

### Troubleshooting Example

```json
{
    "think": "The user has ruled out transmission-side anomalies, so the investigation should shift to the generation branch",
    "ID": "2",
    "response": "Next, the generator operation fault needs to be examined"
}
```

### Solution Example

```json
{
    "think": "The user has confirmed material fatigue, which is a bottom-level fault and can be directly addressed",
    "ID": "23",
    "response": "Material fatigue can be resolved by replacing the impeller assembly"
}
```
\end{promptbox}

\subsection{Prompt for User}
\label{apd:prompt-user}
\begin{promptbox}[Prompt for User]
You are now required to role-play as a user working in the **{field}** domain, collaboratively troubleshooting an occurring fault with the help of an assistant.

Your task is: after receiving a message from the assistant, first determine whether the assistant is currently performing **"Fault Confirmation"** or **"Solution Confirmation"**, and then return the corresponding JSON according to the reply rules.

## Known Fault Information

* **Fault Path (from top level to bottom level)**:
  {error_paths}

* **Bottom-level Faults and Their Corresponding Solutions**:
  {solutions}

## Output Format

Your output must **strictly contain only one JSON object**, in the following format, and **no other content should be included**:

```json
{
  "action": "",
  "name": "",
  "return": "",
  "response": ""
}
```

## Reply Rules

You must first determine the assistant's action type based on its message, then fill in the remaining fields according to the rules for that action. The assistant has two valid action types: **"Fault Confirmation"** and **"Solution Confirmation"**. If the message does not fall into either category, it should be treated as **"Invalid Input."**

---

### **Fault Confirmation**

* If the assistant is asking whether a specific fault exists or requesting you to check the status of a certain node, it is considered **"Fault Confirmation."**

* When the assistant is performing **"Fault Confirmation,"** you must further determine whether the fault exists. Specifically:

  * If the fault name and ID provided by the assistant exactly match a node in the known fault path, set the `"return"` field to `"True"`, indicating that the fault exists.
  * Otherwise, set the `"return"` field to `"False"`, indicating that the fault does not exist.
  * Accurately fill the `"name"` field with the node name that the assistant wants to investigate.
  * The `"response"` field should contain a natural-language reply to the assistant. You must convert your judgment into **vague and non-committal user feedback typical of real production environments**, following these rules:

    * **Do not directly state that a problem has occurred or has not occurred**, and do not use definitive or conclusive expressions.
    * **Components, devices, or modules** involved in the fault should be referred to indirectly via their function, purpose, or appearance.
    * **States, faults, or anomalies** should be replaced with vague sensations, suspected phenomena, observed changes, or possible impacts, without identifying causes, drawing conclusions, or making declarative statements.
    * The rewritten sentence should be natural, conversational, and consistent with the ambiguous expressions commonly used by on-site personnel, while preserving semantic correspondence with the original meaning.

* Examples:

```json
{
  "action": "Fault Confirmation",
  "name": "Chip Overheating",
  "return": "True",
  "response": "I've noticed that the core area responsible for signal processing seems to be giving off more heat than before."
}
```

```json
{
  "action": "Fault Confirmation",
  "name": "Insulation Aging",
  "return": "False",
  "response": "The outer layer used to isolate the circuitry still looks fairly normal for now, and I haven't noticed anything unusual."
}
```

---

### **Solution Confirmation**

* If the assistant provides operational steps or describes how to repair or handle the issue, it is considered **"Solution Confirmation."**

* When the assistant is performing **"Solution Confirmation,"** you must further determine whether the proposed method correctly corresponds to a bottom-level fault and its solution:

  * If the fault and solution both exist in the known fault information and are correctly matched, set `"return"` to `"True"`, fill the `"name"` field with the corresponding bottom-level fault name, and write a natural-language response in the `"response"` field indicating that the issue has been resolved.
  * Otherwise, set `"return"` to `"False"`, fill the `"name"` field with the fault name specified by the user, and write a natural-language response in the `"response"` field indicating that the solution is ineffective.

* Examples:

```json
{
  "action": "Solution Confirmation",
  "name": "Misfire",
  "return": "True",
  "response": "This method is effective and the issue has been resolved."
}
```

```json
{
  "action": "Solution Confirmation",
  "name": "Calibration Pressure Offset",
  "return": "False",
  "response": "The method is ineffective."
}
```

---

### **Invalid Input**

* The following situations are considered **"Invalid Input"**:

  * The assistant's message cannot be associated with either **"Fault Confirmation"** or **"Solution Confirmation,"** such as indicating the investigation is complete or asking whether there are any other issues.
  * The assistant attempts to confirm multiple faults at the same time.
  * The object of solution confirmation is not a bottom-level fault.

* If the assistant's message is judged to be **"Invalid Input,"** set all fields except `"action"` to empty strings.

* Example:

```json
{
  "action": "Invalid Input",
  "name": "",
  "return": "",
  "response": ""
}
```
\end{promptbox}

\subsection{Prompt for User Data Collection}
\label{apd:prompt-user-data}
\begin{promptbox}[Prompt for User Data Collection]
You are an expert at transforming clear, direct technical descriptions into vague, ambiguous, and colloquial on-site observational expressions. Your task is to take the specific sentences I provide and convert the explicit problems, components, or statuses within them into vague, non-standard descriptions.

Rewriting Rules:

* **Do not directly state "a problem occurred" or "no problem occurred"**, and do not use conclusive expressions.
* **Components, equipment, or parts** in the original text can be referred to indirectly by their purpose, function, or visual characteristics.
* **Statuses, faults, or anomalies** in the original text should be replaced with vague impressions, suspected phenomena, observed changes, or potential impacts, without diagnosing the cause, drawing conclusions, or categorizing the issue.
* The rewritten sentences must be natural, conversational, and align with the vague phrasing commonly used by field personnel, while still corresponding to the core meaning of the original sentence.

Examples:

* Original: **Encountered a chip overheating fault**
Obfuscated: **I noticed that the core area responsible for processing signals seems to be giving off more heat than before.**
* Original: **Did not encounter insulation aging**
Obfuscated: **I looked at that outer skin used to isolate the circuits, and it looks pretty normal right now; I haven't noticed any particular changes yet.**

You only need to output the obfuscated on-site phenomenon descriptions. No explanations or additional content are required.

\end{promptbox}

\subsection{Prompt for User Evaluation}
\label{apd:user-eval}
\begin{promptbox}[Prompt for User Evaluation]
You are an **evaluation model** designed to assess the output quality of a *fault-diagnosis user simulation model*.

Your input consists of two parts:

* `query`: a **fault existence check request** from a real production environment
* `response`: the **troubleshooting reply** generated by the user simulation model

Your task is to **objectively score** the quality of the given `response` across multiple dimensions and provide **verifiable justifications**.

---

## Evaluation Dimensions and Rules

### Fuzziness Compliance

**Objective**: Evaluate whether the description aligns with the **vague, non-expert observation style of real users**
**Maximum score: 3**

**Scoring criteria:**

* **0 points**

  * Makes an explicit conclusion

* **1 point**

  * Insufficient fuzziness
  * Contains obvious conclusive or judgmental language
  * But still includes a small number of subjective hedging words (e.g., "feels like")

* **2 points**

  * Generally fuzziness-compliant
  * But some words imply judgment
  * Or clues are overly abstract (e.g., "the state doesn't seem right", "looks kind of weird")

* **3 points (must satisfy all of the following)**

  1. Uses **indirect references** (e.g., "seems", "feels", "kind of", "looks like")
  2. Provides **specific, perceptible clues**
  3. Contains **no state-level or conclusive terms**

**Penalties (-1 point per item, minimum 0):**

* Any of the following word types appear:

  * Direct conclusion terms: *exists / does not exist / normal / fault / broken*
  * State-level judgment terms: *stable, smooth, no issue, something wrong*
  * Range conclusions: *within normal range, not exceeding limits*
* Use of engineering terminology or professional diagnostic language

---

### Naturalness

**Objective**: Evaluate whether the expression resembles **natural speech from real on-site personnel**
**Maximum score: 2**

**Scoring criteria:**

* **0 points**

  * Clearly unnatural
  * Strong machine-like or engineering tone

* **1 point**

  * Semantically clear
  * But template-like or overly formal
  * Or sentence structure is highly similar to responses at other nodes

* **2 points**

  * Fully conversational
  * No template feel
  * Expression does not rely on fixed sentence patterns

**Penalties (-1 point per item, minimum 0):**

* Nearly universal across all nodes
* Obvious boilerplate or fixed phrasing

> **Strong template trigger example:**
> `I checked + X + currently + feels like + state word + temporarily no + abnormality`
>
> Appearance of this structure - naturalness less or equal than 1

---

## Output Format (Strict Requirement)

Output **only JSON**, with no additional explanatory text:

```json
{
  "scores": {
    "fuzziness_compliance": 0,
    "naturalness": 0
  },
  "total_score": 0,
  "comments": ""
}
```

---

## Requirements for the `comments` Field (Mandatory)

* Must explicitly identify:

  * At least **one concrete issue**
  * At least **one concrete strength**
* Vague praise is not allowed
* Must quote **specific wording or structural features** from the `response`
* If the comments mention any of the following issues:

  * "lack of concrete perceptual details"
  * "conclusive wording"
  * "template-like expression"
  * "overly generic description"
    - the corresponding dimension **must be genuinely penalized**

---

## Mandatory Consistency Constraints

1. **No concrete observable clues**
   - fuzziness_compliance cannot be 3

2. **Presence of state-level conclusion terms**
   - fuzziness_compliance less or equal than 2

3. **Template-based sentence structure**
   - naturalness less or equal than 1

4. **Issues mentioned in comments**
   - must be reflected in score deductions

---

## Anchored Examples

### Example A (High Score)

> "I just checked the charging port. After plugging it in, the light didn't turn on immediately - it blinked once after two or three seconds. Feels a bit odd."

* fuzziness_compliance: 3
* naturalness: 2

---

### Example B (Medium Score)

> "I think the charging part seems fairly stable right now, and I didn't notice any issues."

* fuzziness_compliance: 1 ("stable", "didn't notice any issues" = state conclusions)
* naturalness: 2

---

### Example C (Low Score)

> "The charging system is operating normally."

* fuzziness_compliance: 0
* naturalness: 1

---

### Example D (Template Penalty)

> "I checked the system and it currently feels quite stable, with no abnormalities observed so far."

* fuzziness_compliance: 2
* naturalness: 1

---

## Notes

* All scores must strictly fall within the specified ranges
* Do not speculate about components or issues not mentioned in the `query`
* Do not claim problems without deducting points
* Do not use comments to offset actual score deductions

---

## Input

Please score the following sample:

{sample}

Follow the format strictly and output **JSON only**.
\end{promptbox}

\section{Human Experts Details}

Below is the details of human experts as reviewers.
\begin{itemize}
    \item The review was conducted by three trained reviewers, including one senior undergraduate student, one master’s student, and one Ph.D. student, and the review workload was approximately evenly distributed;
    \item The reviewers followed a checklist covering JFTA syntax validity, DAG acyclicity, gate-logic correctness, whether each solution corresponds to a leaf node, and whether the resulting fault paths are interpretable;
    \item The initial generation produced 138 candidate fault trees. After filtering out unsuitable cases, such as scenarios with insufficient distinction between branches or too few nodes, we obtained 123 fault trees. The remaining fault trees were then manually revised when necessary. Typical modifications included standardizing node names, correcting inaccurate AND/OR semantics, and fixing failed LINK references. We rarely changed the core tree structure itself.
\end{itemize}

To facilitate structured inspection and annotation, we also developed a front-end visualization interface tailored to JFTA. This interface allowed reviewers to inspect fault-tree structures more conveniently.

\end{document}